\pdfoutput=1
\documentclass[10pt,twocolumn, twoside]{IEEEtran}
\ifCLASSOPTIONcompsoc
\else
\fi
\ifCLASSINFOpdf
\else
\fi
\usepackage{graphicx}
\usepackage{times}
\usepackage{epsfig}
\usepackage{amsmath}
\usepackage{amssymb}
\usepackage{exscale}
\usepackage[mathscr]{eucal}
\usepackage{bm}
\usepackage{eqlist} % Makes for a nice list of symbols.
\usepackage{flushend}
\usepackage{lettrine}%  dropped capital letter at beginning of paragraph
\usepackage{algorithm}
\usepackage{algorithmic}
\usepackage{url}
\usepackage{multirow} %for table
\usepackage{booktabs} %for table
\usepackage{colortbl} %for color table
\usepackage{lscape}
\usepackage{times} % assumes new font selection scheme installed
\usepackage{amsfonts}
\usepackage{psfrag}
\usepackage{latexsym}
\usepackage{eucal}
\usepackage{caption}
\usepackage{subcaption}
\usepackage{algorithmic}
\usepackage{algorithm}

\usepackage{color}

 \newtheorem{lem}{Lemma}[section]
 \newtheorem{prop}{Proposition}[section]
 \newtheorem{defn}{Definition}[section]

\newcommand{\+}[1]{\bm{#1}}  % for boldface math symboles, \+A makes boldface A, \+{AB} makes boldface AB
\newcommand{\argmin}{\operatornamewithlimits{argmin}}

\newcommand{\Balpha}{\bm{\alpha}} % bold \alpha

\begin{document}

\title{Multimodal Task-Driven Dictionary Learning\\ for Image Classification}

\author{Soheil~Bahrampour,~\IEEEmembership{Member,~IEEE,}
        Nasser~M. Nasrabadi,~\IEEEmembership{Fellow,~IEEE,}
        Asok~Ray,~\IEEEmembership{Fellow,~IEEE,}
        and~W.~Kenneth Jenkins,~\IEEEmembership{Life Fellow,~IEEE}% <-this % stops a space
\IEEEcompsocitemizethanks{\IEEEcompsocthanksitem When this work was achieved, S. Bahrampour, A. Ray, and W. K. Jenkins were with the Department of Electrical Engineering, Pennsylvania State University, University Park, PA 16802, USA; N. M. Nasrabadi was with the Army Research Laboratory, Adelphi, MD 20783. S. Bahrampour is now with Bosch Research and Technology Center, Palo Alto, CA; N. M. Nasrabadi is now with the Computer Science and Electrical Engineering Department at the West Virginia University, WV. \protect
%E-mail: soheil@psu.edu
\IEEEcompsocthanksitem soheil.bahrampour@us.bosch.com, nassernasrabadi@mail.wvu.edu, \hfil\break$\lbrace\textrm{axr2}, \textrm{wkj1} \rbrace\textrm{@psu.edu} $.}% <-this % stops a space
\thanks{}}

%\markboth{Journal of \LaTeX\ Class Files,~Vol.~6, No.~1, January~2007}%
%{Shell \MakeLowercase{\textit{et al.}}: Bare Demo of IEEEtran.cls for Computer Society Journals}

\IEEEcompsoctitleabstractindextext{%
\begin{abstract}
Dictionary learning algorithms have been successfully used for both reconstructive and discriminative tasks, where an input signal is represented with a sparse linear combination of dictionary atoms. While these methods are mostly developed for single-modality scenarios, recent studies have demonstrated the advantages of feature-level fusion based on the joint sparse representation of the multimodal inputs. In this paper, we propose a \textit{multimodal} task-driven dictionary learning algorithm under the joint sparsity constraint (prior) to enforce collaborations among multiple homogeneous/heterogeneous sources of information. In this task-driven formulation, the multimodal dictionaries are learned simultaneously with their corresponding classifiers. The resulting multimodal dictionaries can generate discriminative latent features (sparse codes) from the data that are optimized for a given task such as binary or multiclass classification. Moreover, we present an extension of the proposed formulation using a mixed joint and independent sparsity prior which facilitates more flexible fusion of the modalities at feature level. The efficacy of the proposed algorithms for multimodal classification is illustrated on four different applications -- multimodal face recognition, multi-view face recognition, multi-view action recognition, and multimodal biometric recognition. It is also shown that, compared to the counterpart reconstructive-based dictionary learning algorithms, the task-driven formulations are more computationally efficient in the sense that they can be equipped with more compact dictionaries and still achieve superior performance.
\end{abstract}
}
\maketitle

% is not selected <OR> if conference mode is selected - because compsoc
% conference papers position the abstract like regular (non-compsoc)
\IEEEdisplaynotcompsoctitleabstractindextext

% \fi
\IEEEpeerreviewmaketitle

\section{Introduction}
\label{sec:intro}
It is well established that information fusion using multiple sensors can generally result in an improved recognition performance~\cite{HL97}. It provides a framework to combine local information from different perspectives which is more tolerant to the errors of individual sources~\cite{V00, WSSY02}. Fusion methods for classification are generally categorized into feature fusion~\cite{RG05} and classifier fusion~\cite{RG00} algorithms. Feature fusion methods aggregate extracted features from different sources into a single feature set which is then used for classification. On the other hand, classifier fusions algorithms combine decisions from individual classifiers, each of which is trained using separate sources. While classifier fusion is a well-studied topic, fewer studies have been done for feature fusion, mainly due to the incompatibility of the feature sets~\cite{RKBT07}. A naive way of feature fusion is to stack the features into a longer one~\cite{ZNZH11}. However this approach usually suffers from the curse of dimensionality due to the limited number of training samples~\cite{RG05}. Even in scenarios with abundant training samples, concatenation of feature vectors does not take into account the relationship among the different sources and it may contain noisy or redundant data, which degrade the performance of the classifier~\cite{RKBT07}. However, if these limitations are mitigated, feature fusion can potentially result in improved classification performance~\cite{KTR07, RT09}. %It should be noted that classifier fusion methods do not fully utilize multiple information sources for decision making and decision for each individual classifier is made locally without taking the contextual or global information into account.

Sparse representation classification has recently attracted the interest of many researchers in which the input signal is approximated with a linear combination of \textit{a few} dictionary atoms~\cite{WYGSM09} and has been successfully applied to several problems such as robust face recognition~\cite{WYGSM09}, visual tracking~\cite{ML11}, and transient acoustic signal classification~\cite{ZZNH2012b}. In this approach, a structured dictionary is usually constructed by stacking all the training samples from the different classes. The method has also been expanded for efficient feature-level fusion which is usually referred to as multi-task learning~\cite{C98, SPNC13, SMJMHJ14, MSMSDT14}. Among different proposed sparsity constraints (priors), joint sparse representation has shown significant performance improvement in several multi-task learning applications such as target classification, biometric recognitions, and multiview face recognition~\cite{ZZNH2012b, SPNC13, NNT11, YZZW12}. The underlying assumption is that the multimodal test input can be simultaneously represented by a few dictionary atoms, or training samples, from a multimodal dictionary, that represents all the modalities and, therefore, the resulting sparse coefficients should have the same sparsity pattern. %However, the assumption that \textit{all} inputs from different sources share the same sparsity patterns at atom level may be too restrictive. 
However, the dictionary constructed by the collection of the training samples suffer from two limitations. First, as the number of training samples increases, the resulting optimization problem becomes more computationally demanding. Second, the dictionary that is constructed this way is not optimal neither for the reconstructive tasks~\cite{MBPS09} nor the discriminative tasks~\cite{MBZS08}.

Recently it has been shown that \textit{learning the dictionary} can overcome the above limitations  and significantly improve the performance in several applications including image restoration~\cite{MES08}, face recognition~\cite{YZYZ10} and object recognition~\cite{BBLP10, JLD13}. The learned dictionaries are usually more compact and have fewer dictionary atoms than the number of training samples~\cite{AEB06, MBPS10}. Dictionary learning algorithms can generally be categorized into two groups: unsupervised and supervised. Unsupervised dictionary learning algorithms such as the method of optimal direction~\cite{EAH99} and K-SVD~\cite{AEB06} are aimed at finding a dictionary that yields minimum errors when adapted to reconstruction tasks such as signal denoising~\cite{EA06} and image inpainting~\cite{MBPS09}. Although, the unsupervised dictionary learning has also been used for classification~\cite{YZYZ10}, it has been shown that better performance can be achieved by learning the dictionaries that are adapted to an specific task rather than just the data set~\cite{MBP12, YWLSH12}. These methods are called supervised, or \textit{task-driven}, dictionary learning algorithms. For the classification task, for example, it is more meaningful to utilize the labeled data to minimize the misclassification error rather than the reconstruction error~\cite{KW12}. Adding a discriminative term to the reconstruction error and minimizing a trade-off between them has been proposed in several formulations~\cite{MBZS08, JLD13, MBPSZ08, ZL10}. The incoherent dictionary learning algorithm proposed in~\cite{RSS10} is another supervised formulation which trains class-specific dictionaries to minimize atom sharing between different classes and uses sparse representation for classification. In~\cite{YZFZ11}, a Fisher criterion is proposed to learn structured dictionaries such that the sparse coefficients have small within-class and large between-class scatters. While unsupervised dictionary learning can be reformulated as a large scale matrix factorization problem and solved efficiently~\cite{MBPS09}, supervised dictionary learning is usually more difficult to optimize. More recently, it has been shown that better optimization tool can be used to tackle the supervised dictionary learning~\cite{YWLSH12, YYH10}. This is achieved by formulating it as a bilevel optimization problem~\cite{CMS07, BB08}. In particular, a stochastic gradient descent algorithm has been proposed in~\cite{MBP12} which efficiently solves the dictionary learning problem in a unified framework for different tasks, such as classification, nonlinear image mapping, and compressive sensing. 

The majority of the existing dictionary learning algorithms, including the task-driven dictionary learning~\cite{MBP12}, are only applicable to single source of data. In~\cite{ZJ13}, a set of view-specific dictionaries and a common dictionary are learned for the application of multi-view action recognition. The view-specific dictionaries are trained to exploit view-level correspondence while the common dictionary is trained to capture common patterns shared among the different views. The proposed formulation belongs to the class of dictionary learning algorithms that leverages the labeled samples to learn class-specific atoms while minimizing the reconstruction error. Moreover, it cannot be used for fusion of the heterogeneous modalities. In~\cite{MJVMLG07}, a generative multimodal dictionary learning algorithm is proposed to extract typical templates of multimodal features. The templates represent synchronous transient structures between modalities which can be used for localization applications. More recently, a multimodal dictionary learning algorithm with joint sparsity prior is proposed in~\cite{ZWWZL13} for multimodal retrieval where the task is to find relevant samples from other modalities for a given unimodal query. However, the proposed formulation cannot be readily applied for information fusion in which the task is to find label of a given multimodal query. Moreover, the joint sparsity prior is used in~\cite{ZWWZL13} to couple similarly labeled samples within each modality and is not utilized to extract cross-modality information which is essential for information fusion~\cite{ZZNH2012b}. Furthermore, the dictionaries in~\cite{ZWWZL13} are learned to be generative by minimizing the reconstruction error of data across modalities and, therefore, are not necessary optimal for discriminative tasks~\cite{KW12}.

\begin{figure}[!t]
   \centering
         \includegraphics[scale=0.3]{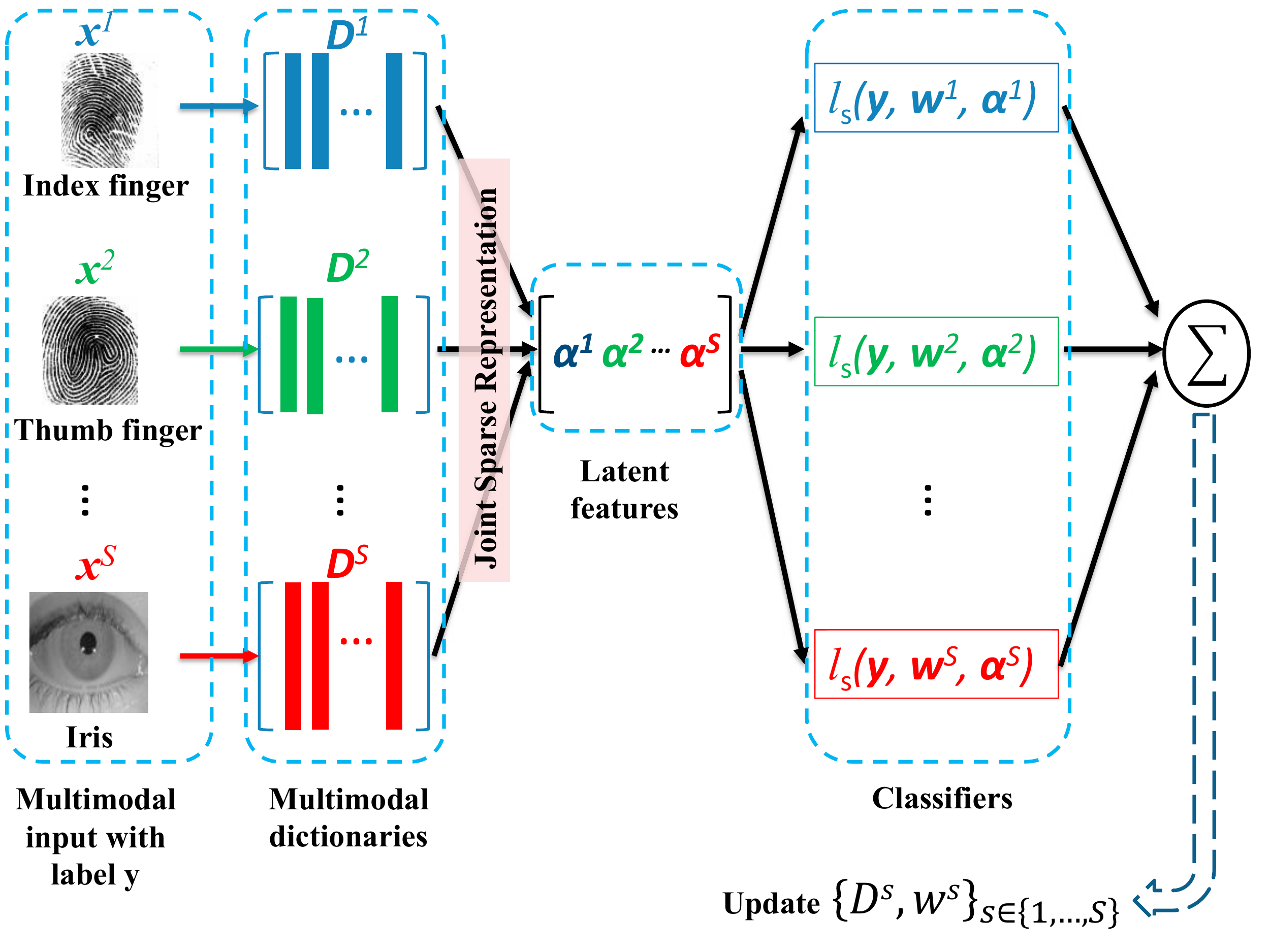}
         \caption{Multimodal task-driven dictionary learning scheme.}
      \label{fig:SMDL}
\end{figure}

This paper focuses on learning discriminative multimodal dictionaries. The major contributions of the paper are as follows: 

\begin{itemize}

\item \textit{Formulation of the multimodal dictionary learning algorithms}: A multimodal task-driven dictionary learning algorithm is proposed for classification using homogeneous or heterogeneous sources of information. Information from different modalities are fused both at the feature level, by using the joint sparse representation, and at the decision level, by combining the scores of the modal-based classifiers. The proposed formulation simultaneously trains the multimodal dictionaries and classifiers under the joint sparsity prior in order to enforce collaborations among the modalities and obtain the latent sparse codes as the optimized features for different tasks such as binary and multiclass classification. Fig.~\ref{fig:SMDL} presents an overview of the proposed framework. An unsupervised multimodal dictionary learning algorithm is also presented as a by-product of the supervised version. 
\item \textit{Differentiability of the bi-level optimization problem}: The main difficulty in proposing such a formulation is that the solution of the corresponding joint sparse coding problem is not differentiable with respect to the dictionaries. While the joint sparse coding has a non-smooth cost function, it is shown here that it is locally differentiable and the resulting bi-level optimization for task-driven multimodal dictionary learning is smooth and can be solved using a stochastic gradient descent algorithm. \footnote{The source code of the proposed algorithm is released here: \url{ https://github.com/soheilb/multimodal_dictionary_learning}}  
\item \textit{Flexible feature-level fusion}: An extension of the proposed framework is presented which facilitates more flexible fusion of the modalities at the feature level by allowing the modalities to have different sparsity patterns. This extension provides a framework to tune the trade-off between independent sparse representation and joint sparse representation among the modalities.
\textit{Improved performance for multimodal classification}: The proposed methods achieve the state-of-the-art performance in a range of different multi-modal classification tasks. In particular, we have provided extensive performance comparison between the proposed algorithms and some of the competing methods from literature for four different tasks of multimodal face recognition, multi-view face recognition, multimodal biometric recognition, and multi-view action recognition. The experimental results on these datasets have demonstrated the usefulness of the proposed formulation, showing that the proposed algorithm can be readily applied to several different application domains.
\item \textit{Improved efficiency for sparse-representation based classification}:
 It is shown here that, compared to the counterpart sparse representation classification algorithms, the proposed algorithms are more computationally efficient in the sense that they can be equipped with more compact dictionaries and still achieve superior performance.

\end{itemize}

\subsection{Paper organization}\label{ssec:PaperOrgan}
The rest of the paper is organized as follows. In Section~\ref{sec:RelatedWorks}, unsupervised and supervised dictionary learning algorithms for single source of information are reviewed. Joint sparse representation for multimodal classification is also reviewed in this section. Section~\ref{sec:MulModDicLea} proposes the task-driven multimodal dictionary learning algorithms. Comparative studies on several benchmarks and concluding results are presented in Section~\ref{sec:Results} and Section~\ref{sec:Conclusions}, respectively.

\subsection{Notation}\label{ssec:Notation}
Vectors are denoted by bold lower case letters and matrices by bold upper case letters. For a given vector $\+x$, $x_i$ is its $i^{th}$ element. For a given finite set of indices $\gamma$, $\+x_{\gamma}$ is the vector formed with those elements of $\+x$ indexed in $\gamma$. Symbol $\rightarrow$ is used to distinguish the row vectors from column vectors, i.e. for a given matrix $\+X$, the $i^{th}$ row and $j^{th}$ column of matrix are represented as $\+x_{i\rightarrow}$ and $\+x_j$, respectively. For a given finite set of indices $\gamma$, $\+X_{\gamma}$ is the matrix formed with those columns of $\+X$ indexed in $\gamma$ and $\+X_{\gamma\rightarrow}$ is the matrix formed with those rows of $\+X$ indexed in $\gamma$. Similarly, for given finite sets of indices $\gamma$ and $\psi$, $\+X_{\gamma\rightarrow, \psi}$ is the matrix formed with those rows and columns of $\+X$ indexed in $\gamma$ and $\psi$, respectively. $x_{ij}$ is the element of $\+X$ at row $i$ and column $j$. The $l_q$ norm, $q \geq 1$, of a vector $\+x \in \mathbb{R}^{m}$ is defined as $\Vert \+x \Vert_{\ell_q} = (\sum_{j=1}^m \vert x_j \vert^q)^{1/q}$. The Frobenius norm and $\ell_{1q}$ norm, $q \geq 1$, of matrix $\+X \in \mathbb{R}^{m \times n}$ is defined as $\Vert \+X \Vert_{F} = \left(\sum_{i=1}^m \sum_{j=1}^n x_{ij}^2\right)^{1/2} $ and $\Vert \+X \Vert_{\ell_{1q}} = \sum_{i=1}^m \Vert \+x_{i\rightarrow} \Vert_{\ell_q}$, respectively. The collection $\lbrace\+x^i \vert i \in \gamma\rbrace$ is shortly denoted as $\lbrace \+x^i\rbrace$.

\section{Background}
\label{sec:RelatedWorks}

\subsection{Dictionary learning}
\label{ssec:DicLea}
Dictionary learning has been widely used in various tasks such as reconstruction, classification, and compressive sensing~\cite{MBP12, ZL10, ZZH13a, OF97}. In contrast to principal component analysis (PCA) and its variants, dictionary learning algorithms generally do not impose orthogonality condition and are more flexible allowing to be well-tuned to the training data. Let $\+X = \left[ \+x_{1}, \+x_{2}, \dots, \+x_{N} \right] \in \mathbb{R}^{n \times N}$ be the collection of $N$ (normalized) training samples that are assumed to be statistically independent. Dictionary $\+D \in \mathbb{R}^{n \times d}$ can then be obtained as the minimizer of the following empirical cost~\cite{YZYZ10}:
\begin{equation} \label{eq:UnsupDicLea}
g_N\left(\+D\right) \triangleq \frac{1}{N} \sum_{i=1}^N l_u\left(\+x_i, \+D\right) 
\end{equation}
over the regularizing convex set $\mathcal{D} \triangleq \lbrace \+D \in \mathbb{R}^{n \times d}\vert \Vert \+d_{k} \Vert_{\ell_2} \leq 1, \forall k = 1, \dots, d \rbrace$, where $\+d_{k}$ is the $k^{th}$ column, or atom, in the dictionary and the unsupervised loss $l_u$ is defined as
\begin{equation}\label{eq:UnsupCost}
l_u\left(\+x, \+D\right) \triangleq \min_{\Balpha \in \mathbb{R}^{d}} \Vert \+x -\+D\Balpha \Vert_{\ell_2}^2 + \lambda_1\Vert \Balpha\Vert_{\ell_1} + \lambda_2\Vert \Balpha\Vert_{\ell_2}^2,
\end{equation}
which is the optimal value of the sparse coding problem with $\lambda_1$ and $\lambda_2$ being the regularizing parameters. While $\lambda_2$ is usually set to zero to exploit sparsity, using $\lambda_2 >0$ makes the optimization problem in Eq.~(\ref{eq:UnsupCost}) strongly convex resulting in a differentiable cost function~\cite{MBP12}. The index $u$ of $l_u$ is used to emphasize that the above dictionary learning formulation is an unsupervised method. It is well-known that one is often interested in minimizing an expected risk, rather than the perfect minimization of the empirical cost~\cite{BB07}. An efficient online algorithm is proposed in~\cite{MBPS09} to find the dictionary $\+D$ as the minimizer of the following stochastic cost over the convex set $\mathcal{D}$:
\begin{equation} \label{eq:OnlineUnsupDicLea}
g\left(\+D\right) \triangleq \mathrm{E}_{\+x} \left[ l_u\left(\+x, \+D\right)\right],
\end{equation}
where it is assumed that the data $\+x$ is drawn from a finite probability distribution $p(\+x)$ which is usually unknown and $\mathrm{E}_{\+x}\left[.\right]$ is the expectation operator with respect to the distribution $p(\+x)$. %A stationary point of the optimization problem can be efficiently obtained by online optimization algorithms. %It follows from the strong law of large numbers that $g_N\left(\+D\right)$ converges to $g\left(\+D\right)$ almost surely as $N\rightarrow \infty$.

The trained dictionary can then be used to (sparsely) reconstruct the input. The reconstruction error has been shown to be a robust measure for classification tasks~\cite{WYGSM09, BRNJ14}. % in the context of well-studied sparse representation classification~\cite{WYGSM09}. 
Another use of a given trained dictionary is for feature extraction where the sparse code $\Balpha^{\star}(\+x, \+D)$, obtained as a solution of~(\ref{eq:UnsupCost}), is used as a feature vector representing the input signal $\+x$ in the classical expected risk optimization for training a classifier~\cite{MBP12}:
\begin{equation}\label{eq:riskMinimization}
\min_{\+w \in \mathcal{W}} \mathrm{E}_{y,\+x} \left[ l\left(y, \+w, \Balpha^{\star}(\+x, \+D)\right)\right] + \frac{\nu}{2} \Vert \+w \Vert_{\ell_2}^2,
\end{equation}
where $y$ is the ground truth class label associated with the input $\+x$, $\+w$ is model (classifier) parameters, $\nu$ is a regularizing parameter, and $l$ is a convex loss function that measures how well one can predict $y$ given the feature vector $\Balpha^{\star}$ and classifier parameters $\+w$. The expectation $\mathrm{E}_{y,\+x} $ is taken with respect to the probability distribution $p(y,\+x)$ of the labeled data. Note that in Eq.~\ref{eq:riskMinimization}, the dictionary $\+D$ is fixed and independent of the given task and class label $y$. In task-driven dictionary learning, on the other hand, a \textit{supervised} formulation is used which finds the optimal dictionary and classifier parameters jointly by solving the following %(non-smooth) 
optimization problem~\cite{MBP12}:
\begin{equation}\label{eq:taskDriv}
\min_{\+D \in \mathcal{D},\+w \in \mathcal{W}} \mathrm{E}_{y,\+x} \left[ l_{su}\left(y, \+w, \Balpha^{\star}(\+x, \+D)\right)\right] + \frac{\nu}{2} \Vert \+w \Vert_{\ell_2}^2.
\end{equation}
The index $su$ of convex loss function $l_{su}$ is used to emphasize that the above dictionary learning formulation is supervised. The learned task-driven dictionary has been shown to result in a superior performance compared to the unsupervised setting~\cite{MBP12}. In this setting, the sparse codes are indeed the optimized latent features for the classifier.

\subsection{Multimodal joint sparse representation}
\label{ssec:SupDicLea}
Joint sparse representation provides an efficient tool for feature-level fusion of sources of information~\cite{ZZNH2012b, SPNC13, CREK05}. Let $\mathcal{S} \triangleq \left\lbrace 1, \dots, S\right\rbrace$ be a finite set of available modalities and let $\+x^s \in \mathbb{R}^{n^s} , s \in \mathcal{S},$ be the feature vector for the $s^{th}$ modality. Also let $\+D^s \in \mathbb{R}^{n^s \times d}$ be the corresponding dictionary for the $s^{th}$ modality. For now, it is assumed that the multimodal dictionaries are constructed by collections of the training samples from different modalities, i.e. $j^{th}$ atom of dictionary $\+D^s$ is the $j^{th}$ training sample from the $s^{th}$ modality. Given a multimodal input $\lbrace\+x^s \vert s \in \mathcal{S}\rbrace$, shortly denoted as $\lbrace\+x^s \rbrace$, an optimal sparse matrix $\+A^{\star}\in \mathbb{R}^{d\times S}$ is obtained by solving the following $\ell_{12}$-regularized reconstruction problem: 
\begin{equation} \label{eq:JSRC}
\argmin_{\+A= \left[\Balpha^1 \dots \Balpha^S\right]} \frac{1}{2}\sum_{s=1}^S\Vert \+x^s - \+D^s\Balpha^s\Vert_{\ell_2}^2 +\lambda\Vert \+A \Vert_{\ell_{12}},
\end{equation}
where $\lambda$ is a regularization parameter. Here $\Balpha^s$ is the $s^{th}$-column of $\+A$ which corresponds to the sparse representation for the $s^{th}$ modality. Different algorithms have been proposed to solve the above optimization problem~\cite{T06, R11}. We use the efficient alternating direction method of multipliers (ADMM)~\cite{PB13} to find $\+A^\star$. The $\ell_{12}$ prior encourages row sparsity in $\+A^\star$, i.e. it encourages collaboration among all the modalities by enforcing the same dictionary atoms from different modalities that present the same event, to be used for reconstructing the inputs $\left\lbrace \+x^s\right\rbrace$. An $\ell_{11}$ term can also be added to the above cost function to extend it to a more general framework where sparsity can also be sought within the rows, as will be discussed in Section~\ref{ssec:l1l2+l1}. It has been shown that joint sparse representation can result in a superior performance in fusing multimodal sources of information compared to other information fusion techniques~\cite{BRNJ14}. We are interested in learning multimodal dictionaries under the joint sparsity prior. This has several advantages over a fixed dictionary consisting of training data. Most importantly, it can potentially remove the redundant and noisy information by representing the training data in a more compact form. Also using the supervised formulation, one expects to find dictionaries that are well-adapted to the discriminative tasks.

\section{Multimodal dictionary learning}
\label{sec:MulModDicLea}
In this section, online algorithms for unsupervised and supervised multimodal dictionary learning are proposed.

\subsection{Multimodal unsupervised dictionary learning}
\label{ssec:MulModUnsupDicLea}
Unsupervised multimodal dictionary learning is derived by extending the optimization problem characterized in Eq.~(\ref{eq:OnlineUnsupDicLea}) and using the joint sparse representation of~(\ref{eq:JSRC}) to enforce collaborations among modalities. %Using the same notations introduced in Section~\ref{sec:RelatedWorks}, 
Let the minimum cost $l^{\prime}_u\left(\lbrace\+x^s, \+D^s\rbrace\right)$ of the joint sparse coding be defined as
%\begin{equation} \label{eq:JSRC2}
%\begin{aligned}
%l^{\prime}_u&\left(\lbrace\+x^s, \+D^s\rbrace\right) \triangleq \\ &\min_{\+A} \frac{1}{2}\sum_{s=1}^S\Vert \+x^s - \+D^s\Balpha^s\Vert_{\ell_2}^2 +\lambda_1\Vert \+A \Vert_{\ell_{12}} + \frac{\lambda_2}{2}\Vert \+A \Vert_{F}^2,
%\end{aligned}
%\end{equation}
\begin{equation} \label{eq:JSRC2}
\min_{\+A} \frac{1}{2}\sum_{s=1}^S\Vert \+x^s - \+D^s\Balpha^s\Vert_{\ell_2}^2 +\lambda_1\Vert \+A \Vert_{\ell_{12}} + \frac{\lambda_2}{2}\Vert \+A \Vert_{F}^2,
\end{equation}
where $\lambda_1$ and $\lambda_2$ are the regularizing parameters. The additional Frobenius norm $\Vert . \Vert_{F}$ compared to Eq.~(\ref{eq:JSRC}) guarantees a unique solution for the joint sparse optimization problem. In the special case when $S=1$, optimization~(\ref{eq:JSRC2}) reduces to the well-studied elastic-net optimization~\cite{ZH05}. By natural extension of the optimization problem~(\ref{eq:OnlineUnsupDicLea}), the unsupervised multimodal dictionaries %${\+D^s}^{\star}$
 are obtained by: 
\begin{equation} \label{eq:UnsupMultiModDicLea}
{\+D^s}^{\star} =\argmin_{\+D^s \in \mathcal{D}^s}\mathrm{E}_{\+x^s} \left[ l^{\prime}_u\left(\lbrace\+x^s, \+D^s\rbrace\right)\right], \forall s \in \mathcal{S}, 
\end{equation}
where the convex set $\mathcal{D}^s$ is defined as
\begin{equation} \label{eq:DicConvexSet}
\mathcal{D}^s \triangleq \lbrace \+D \in \mathbb{R}^{n^s \times d} \vert \Vert \+d_{k} \Vert_{\ell_2} \leq 1, \forall k = 1, \dots, d \rbrace.
\end{equation}
It is assumed that data $\+x^s$ is drawn from a finite (unknown) probability distribution $p(\+x^s)$. The above optimization problem can be solved using the classical projected stochastic gradient algorithm~\cite{AE08} which consists of a sequence of updates as follows:
\begin{equation} \label{eq:UpdateUnsupDicLea}
\+D^s \leftarrow \Pi_{\mathcal{D}^s}\left[\+D^s - \rho_t \nabla_{\+D^s}l^{\prime}_u\left(\lbrace\+x^s_t, \+D^s\rbrace\right) \right],
\end{equation}
where $\rho_t$ is the gradient step at time $t$ and $\Pi_{\mathcal{D}}$ is the orthogonal projector onto set $\mathcal{D}$. The algorithm converges to a stationary point for a decreasing sequence of $\rho_t$~\cite{AE08, B10}. A typical choice of $\rho_t$ is shown in the next section. This problem can also be solved using online matrix factorization algorithm~\cite{MBPS10}.  It should be noted that the while the stochastic gradient descent does converge, it is not guaranteed to converge to a global minimum due to the non-convexity of the optimization problem~\cite{MBPS10, BB07}. However, such stationary point is empirically found to be sufficiently good for practical applications~\cite{MES08, EA06}. % For the single modal scenario, online learning algorithm has been shown to be very efficient in finding a stationary point of this optimization problem. (During section II, review online leanring for single modal dictionary and give the extension for multimodal scenario here). It might be too big for a single paper!

\subsection{Multimodal task-driven dictionary learning}
\label{ssec:MulModSupDicLea}
As discussed in Section~\ref{sec:RelatedWorks}, the unsupervised setting does not take into account the label of the training data, and the dictionaries are obtained by minimizing the reconstruction error. However, for classification tasks, the minimum reconstruction error does not necessarily result in discriminative dictionaries. In this section, a multimodal task-driven dictionary learning algorithm is proposed that enforces collaboration among the modalities both at the feature level using joint sparse representation and the decision level using a sum of the decision scores. We propose to learn the dictionaries ${\+D^s}^{\star}, \forall s \in \mathcal{S},$ and the classifier parameters ${\+w^s}^{\star}, \forall s \in \mathcal{S}$, shortly denoted as the set $\lbrace{\+D^s}^{\star}, {\+w^s}^{\star}\rbrace$, jointly as the solution of the following optimization problem:

\begin{equation}\label{eq:multiModtaskDriv}
\min_{\lbrace{\+D^s} \in \mathcal{D}^s, \+w^s \in \mathcal{W}^s\rbrace} f\left( \lbrace \+D^s, \+w^s \rbrace \right)  + \frac{\nu}{2} \sum_{s=1}^S\Vert \+w^s \Vert_{\ell_2}^2,
\end{equation}
where $f$ is defined as the expected cumulative cost:
\begin{equation}\label{eq:multiModtaskDriv2}
f\left( \lbrace \+D^s, \+w^s \rbrace \right) = \mathrm{E} \sum_{s=1}^S l_{su}(y, \+w^s, {\Balpha^s}^{\star}),
\end{equation}
where ${\Balpha^s}^{\star}$ is the $s^{th}$ column of the minimizer $\+A^{\star}(\lbrace\+x^s,\+D^s\rbrace )$ of the optimization problem~(\ref{eq:JSRC2}) and $l_{su}(y, \+w, {\Balpha})$ is a convex loss function that measures how well the classifier parametrized by $\+w$ can predict $y$ by observing ${\Balpha}$. The expectation is taken with respect to the joint probability distribution of the multimodal inputs $\lbrace\+x^s\rbrace$ and label $y$. Note that ${\Balpha^s}^{\star}$ acts as a hidden/latent feature vector, corresponding to the input $\+x^s$, which is generated by the learned discriminative dictionary ${\+D^s}^{\star}$. In general, $l_{su}$ can be chosen as any convex function such that $l_{su}(y,.,.)$ is twice continuously differentiable for all possible values of $y$. A few examples are given below for binary and multiclass classification tasks. %For the classification task where the domain of $\+y$ is a finite set of labels, then $l_s$ can be chosen as any convex function such that $l_s(y,.)$ is twice continuously differentiable for all possible values of $y$. If $\+y$ belongs to a finite-dimensional real vector space, then $l_s$ can be chosen as any twice continuously differentiable loss function. 

\subsubsection{Binary classification}
\label{sssec:BinaryClass}
In a binary classification task where the label $y$ belongs to the set $\lbrace -1, 1\rbrace$, $l_{su}$ can be naturally chosen as the logistic regression loss 
\begin{equation}\label{eq:logisticLoss}
l_{su}(y, \+w, \Balpha^{\star}) = \log(1 + e^{-y\+w^T\Balpha^{\star}}), 
\end{equation}
where $\+w \in \mathbb{R}^{d}$ is the classifier parameters. Once the optimal $\lbrace\+D^s,\+w^s\rbrace$ are obtained, a new multimodal sample $\lbrace\+x^s\rbrace$ is classified according to sign of $\sum_{s=1}^S {\+w^s}^T\Balpha^{\star}$ due to the uniform monotonicity of $\sum_{s=1}^S l_{su}$. For simplicity, the intercept term for the linear model is omitted here, but it can be easily added. One can also use a bilinear model where, instead of a set of vectors $\lbrace\+w^s\rbrace$, a set of matrices $\lbrace\+W^s\rbrace$ are learned and a new multimodal sample is classified according to the sign of $\sum_{s=1}^S {\+x^s}^T {\+W^s}\Balpha^{\star}$. Accordingly, the $\ell_2$-norm regularization of Eq.~(\ref{eq:multiModtaskDriv}) needs to be replaced with the matrix Frobenius norm. The bilinear model is richer than the linear model and can sometimes result in better classification performance but needs more careful training to avoid over-fitting. 

\subsubsection{Multiclass classification}
\label{sssec:MultiClass}
Multiclass classification can be formulated using a collections of (independently learned) binary classifiers in a one-vs-one or one-vs-all setting. Multiclass classification can also be handled in an all-vs-all setting using the softmax regression loss function. In this scheme, the label $y$ belongs to the set $\lbrace 1, \dots, K\rbrace$ and the softmax regression loss is defined as
\begin{equation}\label{eq:MultiClassLogisticLoss}
l_{su}(y, \+W, \Balpha^{\star}) = -\sum_{k=1}^K 1_{\lbrace y = k\rbrace}\log\left( \frac{e^{\+w_k^T\Balpha^{\star}}}{\sum_{l=1}^K e^{\+w_l^T\Balpha^{\star}}}\right), 
\end{equation}
where $\+W = \left[ \+w_1 \dots \+w_K\right]  \in \mathbb{R}^{d \times K}$, and $1_{\lbrace.\rbrace}$ is the indicator function. Once the optimal $\lbrace\+D^s,\+W^s\rbrace$ are obtained, a new multimodal sample $\lbrace\+x^s\rbrace$ is classified as 
\begin{equation}\label{eq:MultiClassLogisticLoss2}
\operatorname{argmax}_{k \in \left\lbrace 1, \dots, K\right\rbrace} \sum_{s=1}^S\left( \frac{e^{{\+w_k^s}^T{\Balpha^s}^{\star}}}{\sum_{l=1}^K e^{{\+w_l^s}^T{\Balpha^s}^{\star}}}\right).
\end{equation}
In yet another all-vs-all setting, the multiclass classification task can be turned into a regression task in which the scaler label $y$ is changed to a binary vector $\+y \in \mathbb{R}^K$, where the $k^{th}$ coordinate corresponding to the label of $\left\lbrace x^s \right\rbrace$ is set to one and the rest of the coordinates are set to zero. In this setting, $l_{su}$ is defined as 
\begin{equation}\label{eq:QuadraticLoss}
l_{su}(\+y, \+W, \Balpha^{\star}) = \frac{1}{2}\Vert \+y - \+W\Balpha^{\star} \Vert_{\ell_2}^2, 
\end{equation}
where $\+W \in \mathbb{R}^{K \times d}$. Having obtained the optimal $\lbrace\+D^s,\+W^s\rbrace$, the test sample $\left\lbrace x^s \right\rbrace$ is then classified as 
\begin{equation}\label{eq:QuadraticLoss2}
\operatorname{argmin}_{k \in \left\lbrace 1, \dots, K\right\rbrace} \sum_{s=1}^S\Vert \+q^k - \+W^s{\Balpha^s}^{\star} \Vert_{\ell_2}^2,
\end{equation}
where $\+q^k$ is a binary vector in which its $k^{th}$ coordinate is one and its remaining coordinates are zero.

In choosing between the one-vs-all setting, in which independent multimodal dictionaries are trained for each class, and the multiclass formulation, in which multimodal dictionaries are shared between classes, a few points should be considered. In the one-vs-all setting, the total number of dictionary atoms is equal to $dSK$ in the $K$-class classification while in the multiclass setting the number is equal to $dS$. It should be noted that in the multiclass setting a larger dictionary is generally required to achieve the same level of performance to capture the variations among all classes. However, it is generally observed that the size of the dictionaries in multiclass setting is not required to grow linearly as the number of classes increases due to atom sharing among the different classes. Another point to consider is that the class-specific dictionaries of the one-vs-all approach are independent and can be obtained in parallel. In this paper, the multiclass formulation is used to allow feature sharing among the classes. %Overall, if the number of classes is large, one would prefer to use the multiclass setting in order to allow feature-sharing among the classes and for a classification problem with a small number of classes, one may want to use the one-vs-all formulation.

\subsection{Optimization}\label{ssec:Optimization}
The main challenge in optimizing~(\ref{eq:multiModtaskDriv}) is the non-differentiability of $\+A^{\star}(\lbrace\+x^s,\+D^s\rbrace )$. However, it can be shown that although the sparse coefficients $\+A^{\star}$ are obtained by solving a non-differentiable optimization problem, the function $f\left( \lbrace \+D^s ,\+w^s \rbrace \right)$, defined in Eq.~(\ref{eq:multiModtaskDriv2}), is differentiable on $\mathcal{D}^1 \times \cdots \mathcal{D}^S \times\mathcal{W}^1\times \cdots \mathcal{W}^S$, and therefore its gradients are computable. To find the gradient of $f$ with respect to $\+D^s$, one can find the optimality condition of the optimization~(\ref{eq:JSRC2}) or use the fixed point differentiation~\cite{YYH10, BB08} and show that $\+A^{\star}$ is differentiable over its non-zero rows. Without loss of generality, we assume that label $y$ admits a finite set of values such as those defined in Eqs.~(\ref{eq:logisticLoss}) and~(\ref{eq:MultiClassLogisticLoss}). The same algorithm can be derived for the scenario when $\+y$ belongs to a compact subset of a finite-dimensional real vector space as in Eq.~(\ref{eq:QuadraticLoss}). A couple of mild assumptions are required to prove the differentiability of $f$ which are direct generalizations of those required for the single modal scenario~\cite{MBP12} and are listed below:

\textit{Assumption (A)}. The multimodal data $\left( y, \lbrace\+x^s \rbrace\right)$ admit a probability density $p$ with compact support. 

\textit{Assumption (B)}. For all possible values of $y$, $p(y,.)$ is continuous and $l_{su}(y,.)$ is twice continuously differentiable. 

The first assumption is reasonable when dealing with the signal/image processing applications where the acquired values obtained by the sensors are bounded. Also all the given examples for $l_{su}$ in the previous section satisfy the second assumption. Before stating the main proposition of this paper below, the term \textit{active set} is defined.  
 
\defn[Active set] The active set $\Lambda$ of the solution $\+A^{\star}$ of the joint sparse coding problem~(\ref{eq:JSRC2}) is defined to be
\begin{equation}\label{eq:ActSet}
\Lambda = \lbrace j \in \lbrace 1, \dots, d\rbrace: \Vert \+a_{j\rightarrow}^{\star} \Vert_{\ell_2} \neq 0 \rbrace,
\end{equation}
where $\+a_{j\rightarrow}^{\star}$ is the $j^{th}$ row of $\+A^{\star}$.

\prop[Differentiability and gradients of $f$]\label{prop:MainProp}
Let $\lambda_2 > 0$ and the assumptions (\textit{A}) and (\textit{B}) hold. Let $\Upsilon = \cup_{j \in \Lambda} \Upsilon_j $ where $\Upsilon_j = \lbrace j, j+d, \dots, j+(S-1)d\rbrace$. Let the matrix $\+{\hat{D}} \in \mathbb{R}^{n\times |\Upsilon|}$ be defined as 
\begin{equation}\label{eq:Dhat}
\+{\hat{D}} = \left[\+{\hat{D}}_1 \dots \+{\hat{D}}_{|\Lambda |} \right], 
\end{equation} 
where $\+{\hat{D}}_j = \operatorname{blkdiag}(\+d^1_j, \dots, \+d^S_j) \in \mathbb{R}^{n \times S}, \forall j \in \Lambda,$ is the collection of the $j^{th}$ active atoms of the multimodal dictionaries, $\+d^s_j$ is the $j^{th}$ active atom of $\+D^s$, $\operatorname{blkdiag}$ is the block diagonalization operator, and $n = \sum_{s \in \mathcal{S}}n^s$. Also let matrix $\+{\Delta} \in \mathbb{R}^{|\Upsilon| \times |\Upsilon|}$ be defined as
\begin{equation}\label{eq:Delta}
\+{\Delta} = \operatorname{blkdiag}(\+{\Delta}_1, \dots, \+{\Delta}_{|\Lambda |}),
\end{equation}
where $\+{\Delta}_j = \frac{1}{\Vert \+a_{j\rightarrow}^{\star} \Vert_{\ell_2}}\+I - \frac{1}{{\Vert \+a_{j\rightarrow}^{\star} \Vert_{\ell_2}}^3}{\+a_{j\rightarrow}^{\star}}^T\+a_{j\rightarrow}^{\star} \in \mathbb{R}^{S \times S}, \forall j \in \Lambda$, and $\+I$ is the identity matrix.
Then, the function $f$ defined in Eq.~(\ref{eq:multiModtaskDriv2}) is differentiable and $\forall s \in \mathcal{S}$,
\begin{equation}\label{eq:Prop1}
\begin{aligned}
	  \begin{array}{l}
	   \nabla_{\+w^s}f = \mathrm{E} \left[ \nabla_{\+w^s} l_{su}\left(y, \+w^s, {\Balpha^s}^{\star}\right)\right],\\ 
	   \nabla_{\+D^s}f = \mathrm{E} \left[  \left(\+x^s-\+D^s{\Balpha^s}^\star \right)\+{\beta}_{\tilde{s}}^T -\+D^s\+{\beta}_{\tilde{s}}{{\Balpha^s}^\star}^T\right],
	\end{array}
	 %,,
\end{aligned}
\end{equation}
where $\tilde{s} = \left\lbrace s, s+ S, \dots, s+(d-1)S\right\rbrace$ and $\+{\beta} \in \mathbb{R}^{dS}$ is defined as 
\begin{equation}\label{eq:beta}
\+{\beta}_{\Upsilon^c} = \+0, \+{\beta}_{\Upsilon} = ( {\+{\hat{D}}}^T \+{\hat{D}} + \lambda_1\+{\Delta} + \lambda_2\+{I})^{-1}\+g,
\end{equation}
in which $\+g= \operatorname{vec}(\nabla_{{\+A^{\star}_{\Lambda\rightarrow}}^T}\sum_{s=1}^Sl_{su}(y, \+w^s, {\Balpha^s}^{\star}))$, $\Upsilon^c = \left\lbrace 1,\dots, dS\right\rbrace\backslash\Upsilon$, $\+{\beta}_{\Upsilon} \in \mathbb{R}^{|\Upsilon|}$ is formed of  those rows of $\+{\beta}$ indexed by $\Upsilon$, and $\operatorname{vec}(.)$ is the vectorization operator. 

The proof of this proposition is given in the Appendix. A stochastic gradient descent algorithm to find the optimal dictionaries $\lbrace{\+D^s}^{\star}\rbrace$ and classifiers $\lbrace{\+w^s}^\star\rbrace$ is described in Algorithm~\ref{alg:MulModTaskDriDic}. The stochastic gradient descent algorithm is guaranteed to converge under a few assumptions that are mildly stricter than those in this paper (requires three-times differentiability)~\cite{B98}. To further improve the convergence of the proposed stochastic gradient descent algorithm, a classic mini-batch strategy is used in which a small batch of the training data are sampled in each batch, instead of 1 sample, and the parameters are updated using the averaged updates of the batch. This has additional advantage in which ${\+{\hat{D}}}^T \+{\hat{D}}$ and the corresponding factorization of the ADMM for solving the sparse coding problem can be computed once for the whole batch. %However, in practice, stochastic gradient descent has often successfully applied in machine learning algorithms where convergence is not theoretically guaranteed~\cite{MBP12}, as also shown in our experiments.}
For the special case when $S=1$, the proposed algorithm reduces to the single-modal task-driven dictionary learning algorithm in~\cite{MBP12}. Selecting $\lambda_2$ in Eq.~(\ref{eq:JSRC2}) to be strictly positive guarantees the linear equations of~(\ref{eq:beta}) to have a unique solution. In other words, it is easy to show that the matrix $( {\+{\hat{D}}}^T \+{\hat{D}} + \lambda_1\+{\Delta} + \lambda_2\+{I})$ is positive definite given $\lambda_1 \geq 0, \lambda_2 > 0$. However, in practice it is observed that the solution of the joint sparse representation problem is numerically stable since ${\+{\hat{D}}}$ becomes full-column rank when sparsity is sought with a sufficiently large $\lambda_1$, and $\lambda_2$ can be set to zero. It should be noted that the assumption of ${\+{\hat{D}}}$ being a full column rank matrix is a common assumption in sparse linear regression~\cite{MBPS10}. As in any non-convex optimization algorithm, if the algorithm is not initialized properly, it may yield poor performance. Similar to~\cite{MBP12}, the dictionaries $\left\lbrace \+D^s\right\rbrace $ are initialized by the solution of the unsupervised multimodal dictionary learning algorithm. Upon assignment of the initial dictionaries, parameters $\lbrace\+w^s\rbrace$ of the classifiers are set by solving~(\ref{eq:multiModtaskDriv}) only with respect to $\lbrace\+w^s\rbrace$ which is a convex optimization problem. 

\begin{algorithm}[!t]
\footnotesize %\small
\caption{\small Stochastic gradient descent algorithm for multimodal task-driven dictionary learning.}\label{alg:MulModTaskDriDic}
\begin{algorithmic}[1]
	\REQUIRE Regularization parameters $\lambda_1, \lambda_2, \nu$, learning rate parameters $\rho, t_0$, number of iterations $T$, initial dictionaries $\lbrace\+D^s \in \mathcal{D}^s\rbrace_{s \in \mathcal{S}}$, initial model parameters $\lbrace\+w^s \in \mathcal{W}^s\rbrace_{s \in \mathcal{S}}$.
	\ENSURE Learned $\lbrace\+D^s,\+w^s\rbrace$ 
        \FOR{$t= 1, \dots, T$}
        	\STATE Draw a random sample $(\+x^1_t, \dots, \+x^S_t, {y}^{}_t)$ from the training data.
        	\STATE Find solution $\+A^{\star} = \left[{\Balpha^\star}^1 \dots {\Balpha^\star}^S\right] \in \mathbb{R}^{d\times S}$ of the joint sparse coding problem
        		\[
        			\argmin_{\+A = \left[{\Balpha}^1 \dots {\Balpha}^S\right]} \frac{1}{2}\sum_{s=1}^S\Vert \+x^s_t - \+D^s\Balpha^s\Vert_{\ell_2}^2 +\lambda_1\Vert \+A \Vert_{\ell_{12}}+ \frac{\lambda_2}{2}\Vert \+A \Vert_{F}^2.
        		\]
        	\STATE Compute set of active rows $\Lambda$ of $\+A^{\star}$ using~(\ref{eq:ActSet}).
        	\STATE Compute $\+{\hat{D}} \in \mathbb{R}^{n\times |\Upsilon|}$ using~(\ref{eq:Dhat}).
        	\STATE Compute $\+{\Delta} \in \mathbb{R}^{|\Upsilon| \times |\Upsilon|}$ using~(\ref{eq:Delta}).        	
        	\STATE Compute $\+{\beta} \in \mathbb{R}^{dS}$ as:
\begin{equation*}
\+{\beta}_{\Upsilon^c} = \+0, \+{\beta}_{\Upsilon} = ( {\+{\hat{D}}}^T \+{\hat{D}} + \lambda_1\+{\Delta} + \lambda_2\+{I})^{-1}\+g,
\end{equation*}
where $\Upsilon = \cup_{j \in \Lambda} \lbrace j, j+d, \dots, j+(S-1)d\rbrace $ and $\+g = \operatorname{vec}(\nabla_{{\+A^{\star}_{\Lambda\rightarrow}}^T}\sum_{s=1}^Sl_{su}(y^{}_t, \+w^s, {\Balpha^s}^{\star}))$.
        	\STATE Choose the learning rate $\rho^{}_t \leftarrow \min(\rho, \rho\frac{t_0}{t})$.
        	\STATE Update the parameters by a projected gradient step:
        	    \begin{equation*}
        		\begin{aligned}
&\+w^s \leftarrow \Pi_{\mathcal{W}^s}\left[ \+w^s -\rho^{}_t\left( \nabla_{\+w^s}l_{su}\left(y^{}_t, \+w^s, {\Balpha^s}^{\star}\right) + \nu \+w^s \right) \right], \\
& \+D^s \leftarrow \Pi_{\mathcal{D}^s}\left[ \+D^s - \rho^{}_t\left( \left(\+x^s_t-\+D^s{\Balpha^s}^\star \right)\+{\beta}_{\tilde{s}}^T -\+D^s\+{\beta}_{\tilde{s}}{{\Balpha^s}^\star}^T\right) \right], 
\end{aligned}
\end{equation*}
$\forall s \in \mathcal{S}$, where $\tilde{s} = \left\lbrace s, s+ S, \dots, s+(d-1)S\right\rbrace $.
        \ENDFOR
\end{algorithmic}
\end{algorithm}

\subsection{Extension}\label{ssec:l1l2+l1}
We now present an extension of the proposed algorithm with a more flexible structure on the sparse codes. Joint sparse representation relies on the fact that all the modalities share the same sparsity pattern in which, if a multimodal training sample is selected to reconstruct the input, then all the modalities within that training sample are active. However, this group sparsity constraint, imposed by the $\ell_{12}$ norm, may be too stringent for some applications~\cite{BRNJ14, SRSE11}, for example in the scenarios where the modalities have different noise levels or when the heterogeneity of the modalities imposes different sparsity levels for the reconstruction task. A natural relaxation to the joint sparsity prior is to let the multimodal inputs not share the full active set which can be achieved by replacing the $\ell_{12}$ norm with a combination of the $\ell_{12}$ and $\ell_{11}$ norms ($\ell_{12}-\ell_{11}$ norm). Following the same formulation as in Section~\ref{ssec:MulModSupDicLea}, let $\+A^{\star}(\lbrace\+x^s,\+D^s\rbrace )$ in Eq.~(\ref{eq:multiModtaskDriv}) be the minimizer of the following optimization problem:
\begin{equation} \label{eq:JSRC+l1}
\begin{aligned}
\min_{\+A} &\frac{1}{2}\sum_{s=1}^S\Vert \+x^s - \+D^s\Balpha^s\Vert_{\ell_2}^2 \\ & +\lambda_1\Vert \+A \Vert_{\ell_{12}} + \lambda_1^{\prime}\Vert \+A \Vert_{\ell_{11}} + \frac{\lambda_2}{2}\Vert \+A \Vert_{F}^2,
\end{aligned}
\end{equation}
where $\lambda_1^{\prime}$ is the regularization parameter for the added $\ell_{11}$ norm and other terms are the same as those in Eq.~(\ref{eq:JSRC2}). The selection of $\lambda_1$ and $\lambda_1^{\prime}$ influences the sparsity pattern of $\+A^{\star}$. Intuitively, as $\lambda_1/\lambda_1^{\prime}$ increases, the group constraint becomes dominant and more collaboration is enforced among the modalities. On the other hand, small values of $\lambda_1/\lambda_1^{\prime}$ encourage independent reconstructions across modalities. In the extreme case of $\lambda_1$ being set to zero, the above optimization problem is separable across the modalities. The above formulation brings added flexibility with the cost of one additional design parameter which is obtained in this paper using cross-validation. 

Here we present how the Algorithm~\ref{alg:MulModTaskDriDic} should be modified to solve the supervised multimodal dictionary learning problem under the mixed $\ell_{12}-\ell_{11}$ constraint. The proof for obtaining the algorithm is similar to the one for the $\ell_{12}$ norm and is briefly discussed in the appendix. In Algorithm~\ref{alg:MulModTaskDriDic}, let $\+A^{\star}$ be the solution of the optimization problem~(\ref{eq:JSRC+l1}) and let $\Lambda$ be the set of its active rows. Let $\Psi \subseteq \lbrace 1, \dots, S|\Lambda |\rbrace$ be the set of indices with non-zero entries in $\operatorname{vec}({\+A^{\star}_{\Lambda\rightarrow}}^T)$; i.e. it consists of non-zero entries of the active rows of $\+A^{\star}$. Let $\+{\hat{D}}$, $\+{\Delta}$, and $\+g$ be the same as those defined in algorithm~\ref{alg:MulModTaskDriDic}. Then, $\+{\beta} \in \mathbb{R}^{dS}$ is updated as
\begin{equation*}
\+{\beta}_{\Upsilon^c} = \+0, \+{\beta}_{\Upsilon} = ( {\+{\hat{D}}_{\Psi}^T} {\+{\hat{D}}}_\Psi^{} + \lambda_1\+{\Delta}_{\Psi\rightarrow, \Psi}^{} + \lambda_2\+{I})^{-1}{\+g}_{\Psi}^{},
\end{equation*}
where $\Upsilon$ is the set of indices with non-zero entries in $\operatorname{vec}(\+A^{\star^T})$ and $\Upsilon^c = \left\lbrace 1,\dots, dS\right\rbrace\backslash\Upsilon$. Note that $\Upsilon$ is defined over the entire matrix $\+A^{\star}$ while $\Psi$ is defined over its active rows.  The rest of the algorithm remains unchanged.  

\section{Results and discussion}
\label{sec:Results}

The performance of the proposed multimodal dictionary learning algorithms are evaluated on the AR face database~\cite{MB98}, the CMU Multi-PIE dataset~\cite{GMCKB10}, the IXMAS action recognition dataset~\cite{WBR07} and the WVU multimodal dataset~\cite{CRH07}. For these algorithms, $l_s$ is chosen to be the quadratic loss of Eq.~(\ref{eq:QuadraticLoss}) to handle the multiclass classification. In our experiments, it is observed that using the multiclass formulation achieves similar classification performance compared to using the logistic loss formulation of Eq.~(\ref{eq:logisticLoss}) in the one-vs-all setting.
Regularization parameters $\lambda_1$ and $\nu$ are selected using cross-validation in the sets $\lbrace 0.01+0.005k \vert k \in \lbrace -3, 3\rbrace\rbrace$ and $\lbrace 10^{-2}, ..., 10^{-9}\rbrace$, respectively. It is observed that when the number of dictionary atoms is kept small compared to the number of training samples, $\nu$ can be arbitrarily set to a small value, e.g. $\nu = 10^{-8}$, for the normalized inputs. When the mixed $\ell_{12}-\ell_{11}$ norm is used, the regularization parameters $\lambda_1$ and $\lambda_1^{\prime}$ are selected by cross-validation in the set $\lbrace 0.0001, 0.0005, 0.001, 0.005, 0.01, 0.05\rbrace$. The parameter $\lambda_2$ is set to zero in most of the experiments except when using the $\ell_{11}$ prior in Section~\ref{ssec:CMUPie} where a small positive value for $\lambda_2$ was required for convergence. The learning parameter $\rho_t$ is selected according to the heuristic proposed in~\cite{MBP12}, i.e. $\rho_t = \min\left( \rho, \rho\frac{t_0}{t} \right)$ where $\rho$ and $t_0$ are constants. This results in a constant learning rate during the first $t_0$ iterations and an annealing strategy of $1/t$ for the rest of the iterations. It is observed that choosing $t_0 = T/10$, where $T$ is the total number of iterations over the whole training set, works well for all of our experiments. Different values of $\rho$ are tried during the first few iterations and the one that results in minimum error on a small validation set is retained. $T$ is set equal to be 20 in all the experiments. We observed empirically that the selection of these parameters is quite robust and small variations in their values do not affect considerably the obtained results. We also used a mini-batch size of 100 in all our experiments. It should also be noted that design parameters for the competitive algorithms are also selected using cross-validation for a fair comparison. % in all our studied applications. 

\subsection{AR face recognition}
\label{ssec:AR}

\begin{figure}[!t]
   \centering
         \includegraphics[scale=0.28]{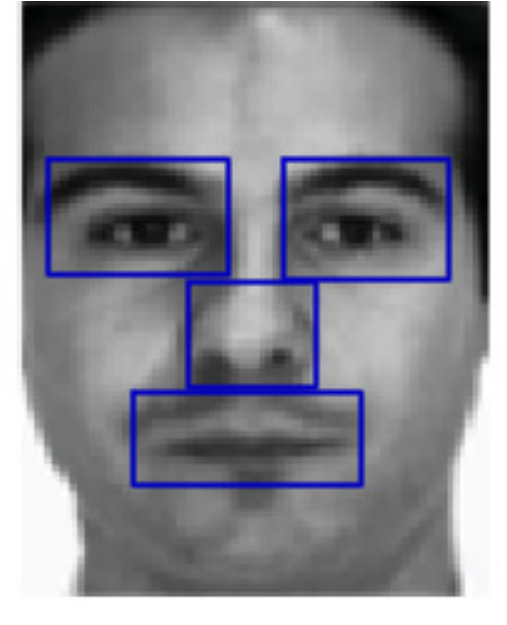}
         \caption{Extracted modalities from a sample in AR dataset.}
      \label{fig:AR}
\end{figure}

\begin{table}[!t] %Table
\caption{Correct classification rates obtained using the whole face modality for the AR database.}
\label{tab:ARIndResults}
\centering
\small
\setlength{\tabcolsep}{2mm}
\begin{tabular}{cccccc}
 SVM & MKL~\cite{RBCG08} & LR & SRC~\cite{WYGSM09} & UDL & SDL~\cite{MBP12} \\
\toprule
  86.43 & 82.86 & 81.00 & 88.86 & 89.58 & \textbf{90.57} \\
\bottomrule
\end{tabular}
\end{table} %¥

\begin{table}[!t] %Table IV
\caption{Comparison of the $\ell_{11}$ and $\ell_{12}$ priors for multimodal classification. Modalities include 1. left periocular, 2. right periocular, 3. nose, 4. mouth, and 5. face.}
\label{tab:l11vsl12}
\centering
\small
\begin{tabular}{ccccc}
Modalities &$\left\lbrace 1,2\right\rbrace $ & $ \left\lbrace 1,2,3\right\rbrace $ & $ \left\lbrace 1,2,3,4\right\rbrace $ & $ \left\lbrace 1,2,3,4,5\right\rbrace $  \\
\toprule
$\textrm{UMDL}_{\ell_{11}}$ & 81.9 & 87.57 & 90.14 & 95.57 \\
$\textrm{UMDL}_{\ell_{12}}$ & 82.6 & 87.86 & 92.00 & 96.29  \\
$\textrm{SMDL}_{\ell_{11}}$ & 83.86 & \textbf{89.86} & 92.42 & 95.86\\
$\textrm{SMDL}_{\ell_{12}}$ & \textbf{86.43} & \textbf{89.86} &  \textbf{93.57} &  \textbf{96.86} \\
\bottomrule
\end{tabular}
\end{table} %¥

\begin{table*}[!t] %Table IV
\caption{Multimodal classification results obtained for the AR datasets}
\label{tab:ARFusResults}
\centering
\small
\begin{tabular}{cccccccccc}
SVM-Maj & SVM-Sum & LR-Maj & LR-Sum & MKL & JSRC~\cite{SPNC13} & JDSRC~\cite{ZNZH11}& $\textrm{SMDL}_{\ell_{11}}$ & $\textrm{SMDL}_{\ell_{12}}$ & $\textrm{SMDL}_{\ell_{12}-\ell_{11}}$\\
\toprule
85.57 & 92.14 & 85.00 & 91.14 &  91.14 & 96.14 & 96.14 & 95.86 & 96.86 & \textbf{97.14}\\

\bottomrule
\end{tabular}
\end{table*}

The AR dataset consists of faces under different poses, illumination and expression conditions, captured in two sessions. A set of 100 users are used, each consisting of seven images from the first session as training samples and seven images from the second session as test samples. A small randomly selected portion of the training set, 50 out of 700, is used as validation set for optimizing the design parameters. Fusion is taken on five modalities which are the left and right periocular, nose, mouth, and the whole face modalities, similar to the setup in~\cite{SPNC13, BRNJ14}. A test sample from the AR dataset and the extracted modalities are shown in Fig.~\ref{fig:AR}. Raw pixels are first PCA-transformed and then normalized to have zero mean and unit $l_2$ norm. The dictionary size for the dictionary learning algorithms is chosen to be four per class, resulting in dictionaries of overall 400 atoms.

%\footnote{Authors of~\cite{SPNC13} kindly provided us with extracted regions.}.

\textit{Classification using the whole face modality}:
 The classification results using the whole face modality are shown in Table~\ref{tab:ARIndResults}. The results are obtained using linear support vector machine (SVM)~\cite{B06}, multiple kernel learning (MKL)~\cite{RBCG08}, logistic regression (LR)~\cite{B06}, sparse representation classification (SRC)~\cite{WYGSM09}, and unsupervised and supervised dictionary learning algorithms (UDL and SDL)~\cite{MBP12}. For the MKL algorithm, linear, polynomial, and RBF kernels are used. The UDL and SDL are equipped with the quadratic classifier~(\ref{eq:QuadraticLoss}). The SDL results in the best performance.  

\textit{$\ell_{11}$ vs $\ell_{12}$ sparse priors for multimodal classification}:  A straightforward way of utilizing the single-modal dictionary learning algorithms, namely UDL and SDL, for multimodal classification is to train independent dictionaries and classifiers for each modality and then combine the individual scores for a fused decision. This way of fusion is equivalent to using the $\ell_{11}$ norm on $\+A$, instead of $\ell_{12}$ norm, in Eq.~(\ref{eq:JSRC2}) (or setting $\lambda_1$ to zero in Eq.~(\ref{eq:JSRC+l1})) which does not enforce row sparsity in the sparse coefficients. We denote the corresponding unsupervised and supervised multimodal dictionary learning algorithms using only the $\ell_{11}$ norm as $\textrm{UMDL}_{\ell_{11}}$ and $\textrm{SMDL}_{\ell_{11}}$, respectively. Similarly, the proposed unsupervised and supervised multimodal dictionary learning algorithms using the $\ell_{12}$ norm are denoted as $\textrm{UMDL}_{\ell_{12}}$ and $\textrm{SMDL}_{\ell_{12}}$. Table~\ref{tab:l11vsl12} compares the performance of the multimodal dictionary learning algorithms under the two priors. As shown, the proposed algorithms with $\ell_{12}$ prior, which enforces collaborations among the modalities, have better fusion performances than those with $\ell_{11}$ prior. In particular, $\textrm{SMDL}_{\ell_{12}}$ has significantly better performance than the $\textrm{SMDL}_{\ell_{11}}$ for fusion of the first and second (left and right periocular) modalities. This agrees with the intuition that these modalities are highly correlated and learning the multimodal dictionaries jointly indeed improves the recognition performance.% It should be noted that the objective of this paper is not to present a rigorous comparison on selection of the group sparsity priors but to present a supervised dictionary learning algorithm under the frequently used joint sparsity prior. %which outperforms the corresponding reconstructive approach, as illustrated later.

%\item \textit{Sensitivity to the regularization parameter $\lambda_1$}: To evaluate the sensitivity of the proposed algorithms to the regularization parameter $\lambda_1$, the performance of the $\textrm{SMDL}_{\ell_{12}}$ is studied using a number of different regularization parameters. Fig.~\ref{fig:} shows the classification performance for different fusion settings, i.e. using iris modalities, using fingerpront modalities, and using all available modalities. The results suggests a consistent performance of the $\textrm{SMDL}_{\ell_{12}}$ for a relatively large range of the regularization parameters. The figure suggests that selecting the regularization parameter to be too big or too small can degrade the $\textrm{SMDL}_{\ell_{12}}$ performance. 

\textit{Comparison with other fusion methods}: The performances of the proposed fusion algorithms under different sparsity priors are compared with those of the several state-of-the-art decision-level and feature-level fusion algorithms. In addition to $\ell_{11}$ and $\ell_{12}$ priors, we evaluate the proposed supervised multimodal dictionary learning algorithm with the mixed $\ell_{12}-\ell_{11}$ norm which is denoted as $\textrm{SMDL}_{\ell_{12}-\ell{11}}$. One way to achieve decision-level fusion is to train independent classifiers for each modality and aggregate the outputs by either adding the corresponding scores of each modality to come up with the fused decision, or using the majority voting among the independent decisions obtained from different modalities. These approaches are abbreviated with \textit{Sum} and \textit{Maj}, respectively, and are used with SVM and LR classifiers for decision-level fusion. The proposed methods are also compared with feature-level fusion methods including the joint sparse representation classifier (JSRC)~\cite{SPNC13}, joint dynamic sparse representation classifier (JDSRC)~\cite{ZNZH11}, and MKL. For the JSRC and JDSRC, the dictionary consists of all the training samples. Table~\ref{tab:ARFusResults} compares the performance of our proposed algorithms with the other fusion algorithms for the AR dataset. As expected, the multimodal fusion results in significant performance improvement compared to using only the whole face modality. Moreover, the proposed $\textrm{SMDL}_{\ell_{12}}$ and $\textrm{SMDL}_{\ell_{12}-\ell_{11}}$ achieve the superior performances.
 %Fig.~\ref{fig:ARCMCPlot} shows the corresponding cumulative matched score curves (CMC) for the competitive methods. CMC is a performance measure, similar to ROC, which is originally proposed for biometric recognition systems~\cite{BCPRS05}. 

%\textit{Comparison of different multiclass formulations}:

%\begin{figure}[t]
%        \centering
%        \includegraphics[scale=0.18]{Figures/CMCAR.eps}                
%        \caption{CMC plots obtained by fusing the modalities on the AR dataset.}
%        \label{fig:ARCMCPlot}
%\end{figure}

%\begin{table*}[!t] %Table IV
%\caption{Comparison of the reconstructive (JSRC and JSRC-UDL) and discriminative (SDDL) versions of the joint sparsity prior for multimodal classification obtained on the AR dataset for the different number of dictionary atoms per class.}
%\label{tab:VaryingAtomAR}
%\centering
%\small
%\begin{tabular}{cccccccc}
%Number of atoms per class & $ 1$ & $ 2$ & $ 3$ & $ 4$ & $ 5$ & $ 6$ & $ 7$   \\
%\toprule
%JSRC & 46.14 & 69.00 & 79.57 & 88.14 & 91.00 & 94.43 & \textbf{96.14}\\
%JSRC-UDL & 71.71 & 78.86 & 83.57 & 91.14 & 94.85 & 96.28 & \textbf{96.14} \\
%%SDL-Sum & 91.28 & 94.28 & 94.86 & 96.43 & 94.85 & 96.28 & 96.14 \\
%SMDL & \textbf{91.28}& \textbf{95} & \textbf{95.71} &  \textbf{96.86} &  \textbf{97.14} & \textbf{96.71}& 96.00\\
%\bottomrule
%\end{tabular}
%\end{table*} %¥

\begin{table}[!t] %Table IV
\caption{Comparison of the reconstructive-based (JSRC and JSRC-UDL) and the proposed discriminative-based ($\textrm{SMDL}_{\ell_{12}}$) classification algorithms obtained using the joint sparsity prior for different numbers of dictionary atoms per class on the AR dataset.}
\label{tab:VaryingAtomAR}
\centering
\small
\begin{tabular}{cccc}
atoms/class & JSRC & JSRC-UDL & $\textrm{SMDL}_{\ell_{12}}$  \\
\toprule
$ 1$ & 46.14 & 71.71 & \textbf{91.28} \\
$ 2$ & 69.00 & 78.86 & \textbf{95.00}  \\
$ 3$ & 79.57 & 83.57 & \textbf{95.71} \\
$ 4$ & 88.14 & 91.14 & \textbf{96.86} \\
$ 5$ & 91.00 & 94.85 & \textbf{97.14} \\
$ 6$ & 94.43 & 96.28 & \textbf{96.71} \\
$ 7$ & \textbf{96.14} &\textbf{96.14} & 96.00 \\
\bottomrule
\end{tabular}
\end{table} %¥

\begin{figure}[t]
        \centering
        \includegraphics[scale=0.18]{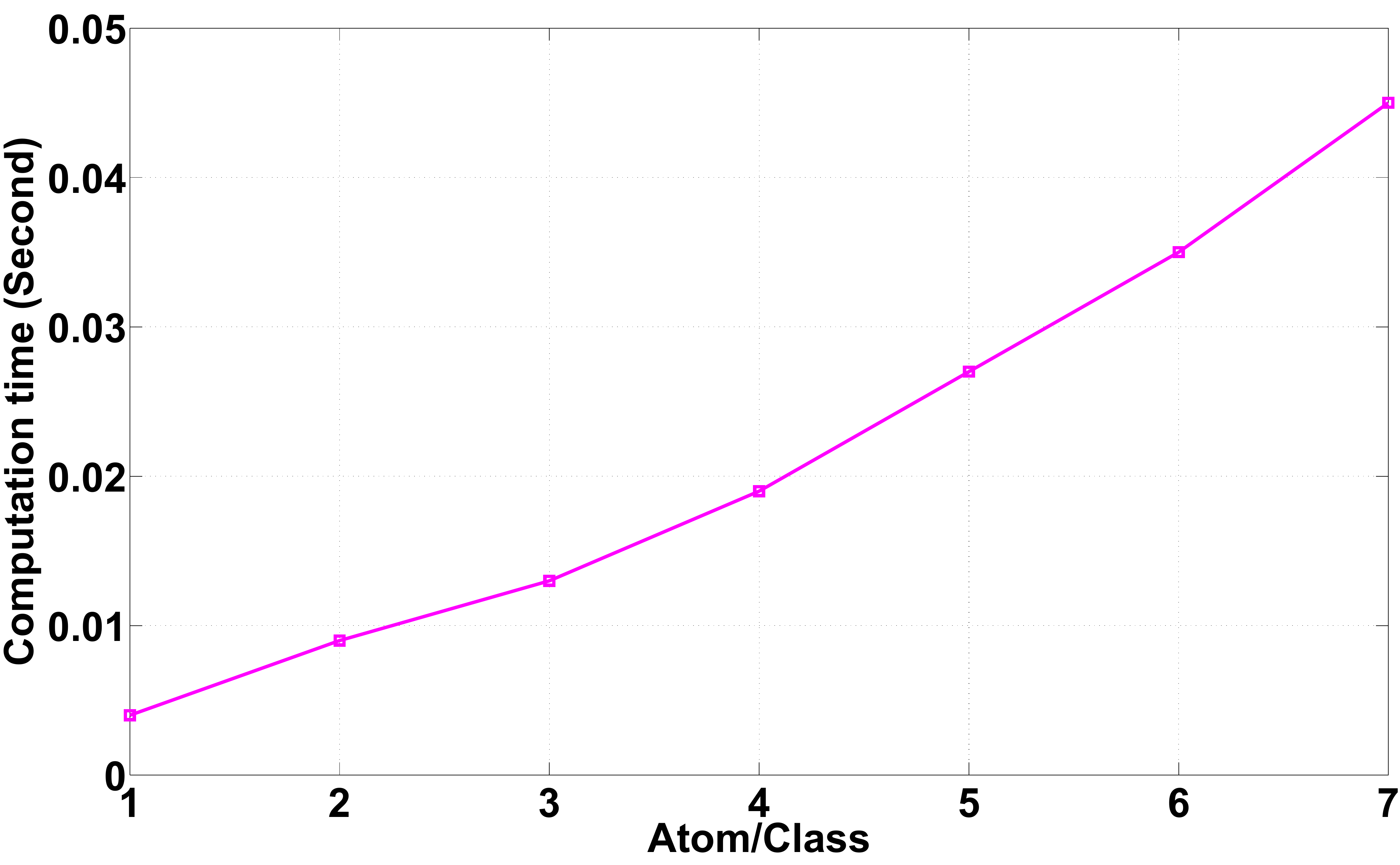}                
        \caption{Computational time required to solve the optimization problem~(\ref{eq:JSRC2}) for a given test sample.}
        \label{fig:ComTime}
\end{figure}

\begin{table}[!t] %Table IV
\caption{Comparison of the supervised multimodal dictionary learning algorithms with different sparsity priors for face recognition under occlusion on the AR dataset.}
\label{tab:AROcclusion}
\centering
\small
\begin{tabular}{ccc}
$\textrm{SMDL}_{\ell_{12}}$ & $\textrm{SMDL}_{\ell_{11}}$ & $\textrm{SMDL}_{\ell_{12}-\ell_{11}}$  \\
\toprule
 89.00 & 90.54  & \textbf{91.15} \\
\bottomrule
\end{tabular}
\end{table} %¥
 
\textit{Reconstructive vs discriminative formulation with joint sparsity prior}: Comparison of the algorithms with joint sparsity priors in Table~\ref{tab:ARFusResults} indicates that the proposed $\textrm{SMDL}_{\ell_{12}}$ algorithm equipped with dictionaries of size 400 achieves relatively better results than the JSRC that uses dictionaries of size 700. The results confirm the idea that by using the supervised formulation, compared to using the reconstruction error, one can achieve better classification performance even with more compact dictionaries. For further comparison, an experiment is performed in which the correct classification rates of the reconsturtive and discriminative formulations are compared when the their dictionary sizes are kept equal. For a given number of dictionary atoms per class $d$, dictionaries of JSRC are thus constructed by random selection of $d$ train samples from different classes. This is different from the standard JSRC, utilized for the results in Table~\ref{tab:ARFusResults}, in which all the training samples are used to construct the dictionaries~\cite{SPNC13}. Moreover, to utilize all the available training samples for the reconstructive approach and make a more meaningful comparison, we use the unsupervised multimodal dictionary learning algorithm of Eq.~(\ref{eq:UnsupMultiModDicLea}) to train class-specific sub-dictionaries which minimizes the reconstruction error in approximating the training samples for a given class. These sub-dictionaries are then stacked to construct the final dictionaries, similar to the approach in~\cite{YZYZ10}. We call this algorithm as JSRC-UDL to indicate that the dictionaries are indeed learned by the reconstructive formulation. Table~\ref{tab:VaryingAtomAR} summarizes the recognition performance of JSRC and JSRC-UDL in comparison to the proposed $\textrm{SMDL}_{\ell_{12}}$, which enjoys a discriminative formulation, for different number of dictionary atoms per class. As seen, $\textrm{SMDL}_{\ell_{12}}$ outperforms the reconstructive approaches, especially when the number of dictionary is chosen to be relatively small. This is the main advantage of $\textrm{SMDL}_{\ell_{12}}$ compared to the reconstructive approaches in which more compact dictionaries can be used for the recognition task that is important for the real-time applications. It is clear that reconstructive model can only result in comparable performance when the dictionary size is chosen to be relatively large. On the other hand, the $\textrm{SMDL}_{\ell_{12}}$ algorithm may get over-fitted with the large number of dictionary atoms. In terms of computational expense at test time, as discussed in~\cite{SPNC13}, the time required to solve the optimization problem~(\ref{eq:JSRC2}) is expected to be linear in the dictionary size using the efficient ADMM if the required matrix factorization is cashed beforehand. Typical computational time to solve~(\ref{eq:JSRC2}) for a given multimodal test sample is shown in Fig.~\ref{fig:ComTime} for different dictionary sizes. As expected, it increases linearly as the size of the dictionary increases. This illustrates the advantage of the $\textrm{SMDL}_{\ell_{12}}$ algorithm that results in the state-of-the-art performance with more compact dictionaries. 

%the JSRC has almost zero computational cost at the training phase and leaves all the computation burden for the test time. On the other hand, $\textrm{SMDL}_{\ell_{12}}$ has a higher computation cost during the training phase, which is usually off-line, but more compact and yet discriminative dictionaries can be used at the test time. 

\textit{Classification in presence of disguise}:
The AR dataset also contains 600 occluded samples per session, overall 1200 images, where the faces are disguised using sun glasses or scarf. Here we use these additional images to evaluate the robustness of the proposed algorithms. Similar to previous experiments, images from session 1 are used as training samples and images from session 2 are used as test data. Classification performance under different sparsity priors are shown in Table~\ref{tab:AROcclusion} and as expected, the $\textrm{SMDL}_{\ell_{12}-\ell_{11}}$ achieves the best performance. In presence of occlusion, some of the modalities are less coupled and the joint sparsity prior among all the modalities may be too stringent as is also reflected in the results.

\subsection{Multi-view recognition}\label{ssec:MultiViewRecog}

\subsubsection{Multi-view face recognition}
\label{ssec:CMUPie}

\begin{figure}[!t]
   \centering
         \includegraphics[scale=0.18]{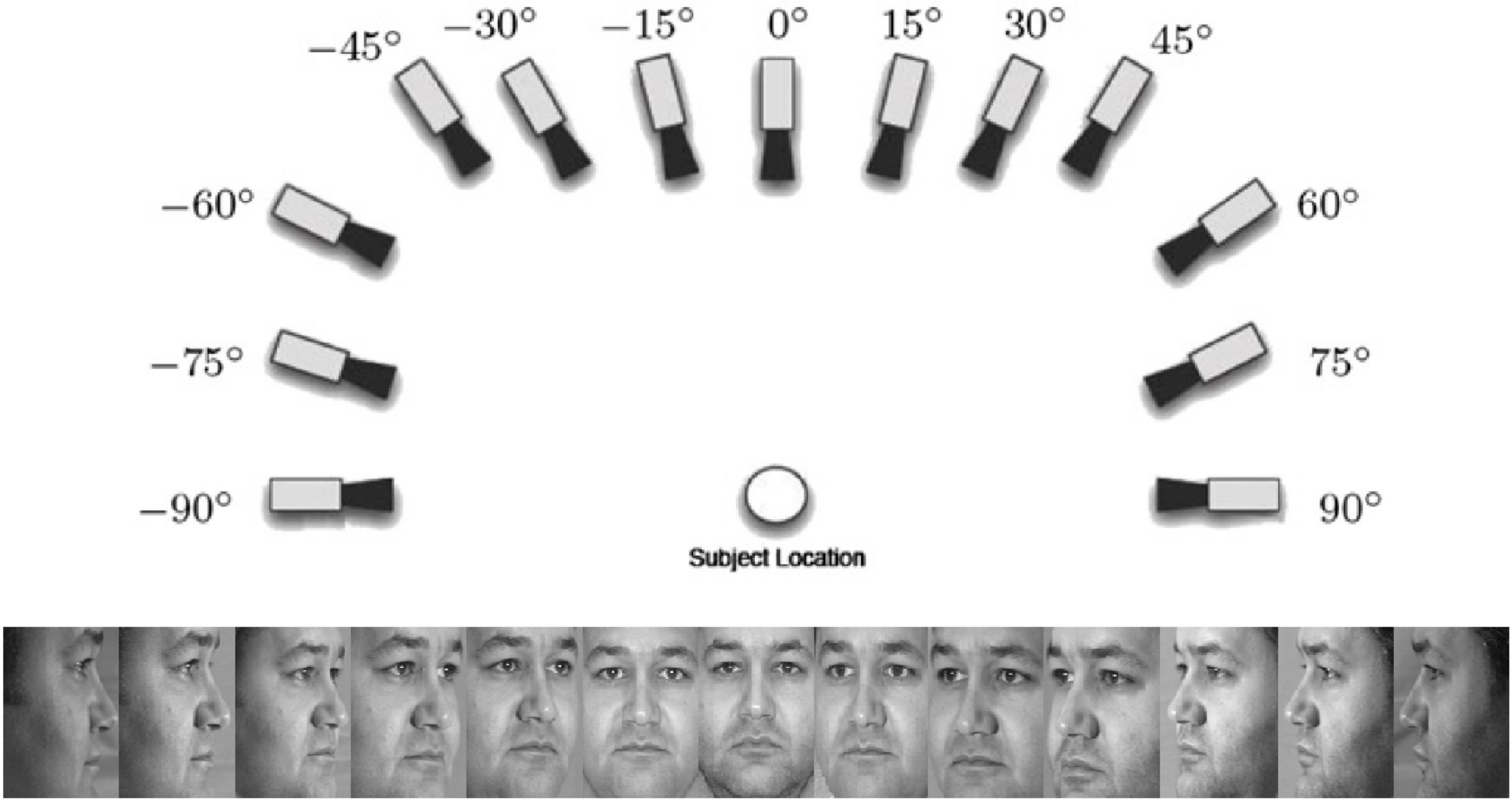}
         \caption{Configurations of the cameras and sample multi-view images from CMU Multi-Pie dataset.}
      \label{fig:Multi-Pie}
\end{figure}

In this section, the performance of the proposed algorithm is evaluated for multi-view face recognition using the CMU Multi-PIE dataset~\cite{GMCKB10}. The dataset consists of a large number of face images under different illuminations, viewpoints, and expressions which are recorded in four sessions over the span of several months. Subjects were imaged using 13 cameras at different view-angles of $\lbrace 0^{\circ}, \pm 15^{\circ}, \pm 30^{\circ}, \pm 45^{\circ}, \pm 60^{\circ}, \pm 75^{\circ}, \pm 90^{\circ}\rbrace$ at head height. Illustrations for the multiple camera configurations, as well as sample multi-view images are shown in Fig.~\ref{fig:Multi-Pie}. We use the multi-view face images for 129 subjects that are present in all sessions. The face regions for all the poses are extracted manually and resized to $10 \times 8$. Similar to the protocol used in~\cite{ZNZH12a}, images from session 1 at views $\lbrace 0^{\circ}, \pm 30^{\circ}, \pm 60^{\circ},  \pm 90^{\circ}\rbrace$ are used as training samples. Test images are obtained from all available view angles from session 2 to have a more realistic scenario in which not all the testing poses are available in the training set. To handle multi-view recognition using the multi-modal formulation, we divide the available views into three sets of $\lbrace -90^{\circ}, -75^{\circ}, -60^{\circ}, -45^{\circ}\rbrace$, $\lbrace -30^{\circ}, -15^{\circ}, 0^{\circ}, 15^{\circ}, 30^{\circ}, \rbrace $, $\lbrace 45^{\circ}, 60^{\circ}, 75^{\circ}, 90^{\circ}\rbrace$, each of which forms a modality. A test sample is then constructed by randomly selecting an image from each modality. Two thousand test samples are generated in this way. The dictionary size for the dictionary learning algorithms is chosen to have two atoms per class.  

The classification results obtained using individual modalities are shown in Table~\ref{tab:CMUPieIndResults}. As expected, better classification performance is obtained using the frontal view. Results of the multi-view face recognition is shown in Table~\ref{tab:MultiPieFusResults}. The proposed supervised dictionary learning algorithms outperform the corresponding unsupervised methods and other fusion algorithms. The $\textrm{SMDL}_{\ell_{12}-\ell_{11}}$ results in the state-of-the-art performance. It is consistently observed in all the studied applications that the multimodal dictionary learning algorithm with the mixed prior results in better performance than those with individual $\ell_{12}$ or $\ell_{12}$ prior. However, it requires one additional regularizing parameter to be tuned. For the rest of the paper, the performance of the proposed dictionary learning algorithms are only reported under the individual priors.

\begin{table}[!t] %Table
\caption{Correct classification rates obtained using individual modalities in the CMU Multi-PIE database.}
\label{tab:CMUPieIndResults}
\centering
\small
\setlength{\tabcolsep}{2mm}
\begin{tabular}{ccccccc}
 View & SVM & MKL & LR & SRC & UDL & SDL \\
\toprule
Left& 47.30 & \textbf{52.85} & 43.65 & 49.85 & 47.80 & 50.45 \\
Frontal & 41.15 & 54.10 & 45.40 & 54.25 & 52.10 & \textbf{56.10}  \\
Right & 47.30 & 51.85 & 42.85 & \textbf{52.55} & 43.10 & 48.50 \\
\bottomrule
\end{tabular}
\end{table} %¥

\begin{table}[!t] 
\caption{Correct classification rates (CCR) obtained using multi-view images on the CMU Multi-PIE database.}
\label{tab:MultiPieFusResults}
\centering
\small
\begin{tabular}{cccc}
Algorithm & CCR & Algorithm & CCR \\ \toprule
SVM-Maj & 62.95 & LR-Maj & 69.40 \\
SVM-Sum & 69.30 & LR-Sum & 71.10\\
MKL & 72.40 & JSRC & 73.30 \\
JDSRC & 70.20 & $\textrm{UMDL}_{\ell_{11}}$ & 74.80\\
$\textrm{SMDL}_{\ell_{11}}$ & 77.25 & $\textrm{UMDL}_{\ell_{12}}$ & 70.50\\
$\textrm{SMDL}_{\ell_{12}}$ & 76.10 & $\textrm{SMDL}_{\ell_{12}-\ell_{11}}$ & \textbf{81.30}\\
\bottomrule
\end{tabular}
\end{table} 

%\begin{table*}[!t] %Table IV
%\caption{Multi-view classification results obtained on the CMU Multi-PIE database.}
%\label{tab:MultiPieFusResults}
%\centering
%\small
%\begin{tabular}{cccccccccc}
%SVM-Sum & LR-Sum & MKL & JSRC & JDSRC & $\textrm{UMDL}_{\ell_{11}}$ & $\textrm{SMDL}_{\ell_{11}}$ & $\textrm{UMDL}_{\ell_{12}}$ & $\textrm{SMDL}_{\ell_{12}}$ & $\textrm{SMDL}_{\ell_{12}-\ell_{11}}$ \\
%\toprule
%69.30 & 71.10 & 72.40 & 73.30 & 70.20  & 74.80 & 77.25  &  70.50 & 76.10 & \textbf{81.30}\\
%\bottomrule
%\end{tabular}
%\end{table*} %¥

\subsubsection{Multi-view action recognition}
\label{ssec:ixmas}
\begin{figure}[!t]
   \centering
         \includegraphics[scale=0.33]{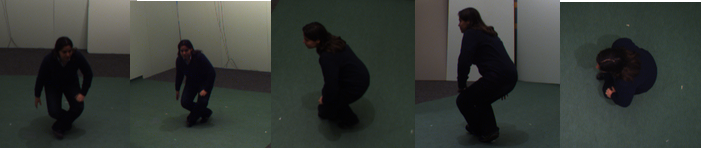}
         \caption{Sample frames of the IXMAS dataset from 5 different views. }
      \label{fig:ixmas}
\end{figure}

This section presents the results of the proposed algorithm for the purpose of multi-view action recognition using the IXMAS dataset~\cite{WBR07}. Each action is recorded simultaneously by cameras from five different viewpoints, which are considered as modalities in this experiment. A multimodal sample of the IXMAS dataset is shown in Fig.~\ref{fig:ixmas}. The dataset contains 11 action classes where each action is repeated three times by each of the ten actors, resulting in 330 sequences per view. The dataset include actions such as check watch, cross arms, and scratch head. Similar to the work in~\cite{WBR07, TS08, WKSL13}, leave-one-actor-out cross-validation is performed and samples from all five views are used for training and testing. 

We use dense trajectories as features which are generated using the publicly available code~\cite{WKSL13} in which a 2000 word codebook is generated by a random subset of these trajectories and the k-means clustering as in~\cite{GSLW14}. Note that Wang et al.~\cite{WKSL13} used HOG, HOF, and MBH descriptors in addition to the dense trajectories. However, here only dense trajectory descriptors are used. The number of dictionary atoms for the proposed dictionary learning algorithms are chosen to be 4 atoms per class, resulting in a dictionary of 44 atoms per view. The five dictionaries for JSRC are constructed using all the training samples, thus each dictionary, corresponding to a different view, has 297 atoms. 

\begin{figure}[!t]
   \centering
         \includegraphics[scale=0.44]{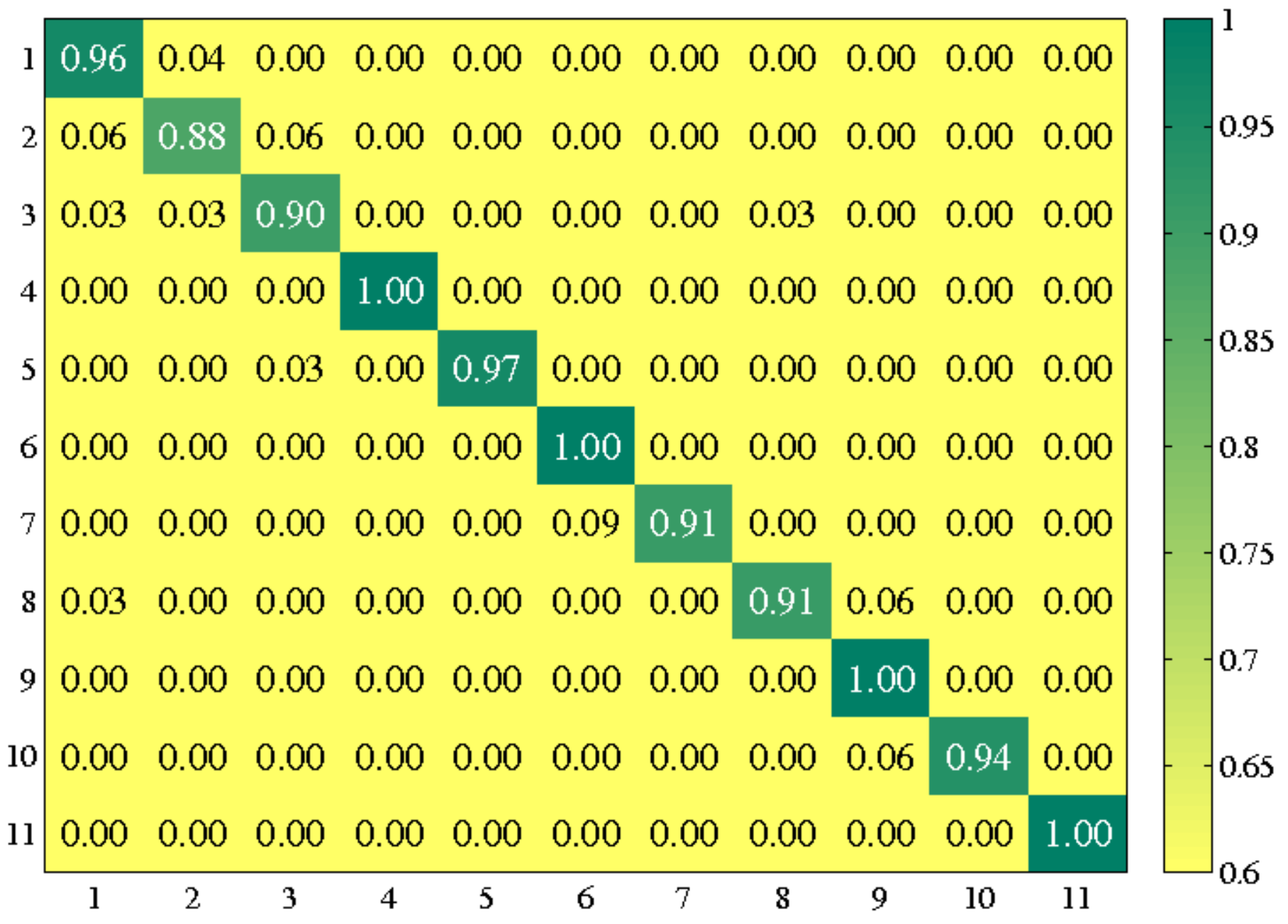}
         \caption{The confusion matrix obtained by the $\textrm{SMDL}_{\ell_{12}}$  algorithm on the IXMAS dataset. The actions are 1: check watch, 2: cross arms, 3: scratch head, 4: sit down, 5: get up, 6: turn around, 7: walk, 8: wave, 9: punch, 10: kick and 11: pick up.}
      \label{fig:ixmas2}
\end{figure}

Table~\ref{tab:ixmasResults} shows average accuracies over all classes obtained using the existing algorithms and the state of the art algorithms. The Wang et al. 1~\cite{WKSL13} algorithm uses only the dense trajectories as feature, similar to our setup. The Wang et al. 2~\cite{WKSL13} algorithm, however, uses HOG, HOF, MBH descriptors and the spatio-temporal pyramids in addition to the trajectory descriptor. The results show that the proposed $\textrm{SMDL}_{\ell_{12}}$ algorithm achieves the superior performance while the $\textrm{SMDL}_{\ell_{11}}$ algorithm achieves the second best performance. This indicates that sparse coefficients generated by the trained dictionaries are indeed more discriminative than the engineered features. The resulting confusion matrix of the $\textrm{SMDL}_{\ell_{12}}$ algorithm is shown in Fig.~\ref{fig:ixmas2}.

\begin{table}[!t] 
\caption{Correct classification rates (CCR) obtained for multi-view action recognition on the IXMAS database.}
\label{tab:ixmasResults}
\centering
\small
\begin{tabular}{cccc}
Algorithm & CCR & Algorithm & CCR \\ \toprule
Junejo et al.~\cite{JDLP11} & 79.6 &  Tran and Sorokin~\cite{TS08} & 80.2 \\
Wu et al.~\cite{WXDL11} & 88.2 & Wang et al. 1~\cite{WKSL13}  & 87.8\\
Wang et al. 2~\cite{WKSL13} & 93.6 & JSRC &  93.6\\
$\textrm{UMDL}_{\ell_{11}}$ &  90.3 & $\textrm{SMDL}_{\ell_{11}}$ & 93.9\\
$\textrm{UMDL}_{\ell_{12}}$ & 90.6 & $\textrm{SMDL}_{\ell_{12}}$ & \textbf{94.8}\\
\bottomrule
\end{tabular}
\end{table} 

\subsection{Multimodal biometric recognition}
\label{ssec:WVU}

\begin{table*}[!t] %Table
\caption{Correct classification rates obtained using individual modalities in the WVU database.}
\label{tab:WVUIndResults}
\centering
\small
\setlength{\tabcolsep}{2mm}
\begin{tabular}{ccccccc}
 & Finger 1 & Finger 2 & Finger 3 & Finger 4 & Iris 1 & Iris 2 \\
\toprule
SVM & $56.77 \pm 0.72$ & $82.95 \pm 2.15$ & $55.83 \pm 2.03$ & $80.47 \pm 0.91$ & $60.67 \pm 1.78$ & $57.52 \pm 1.95$\\
MKL & $61.81 \pm 1.39$ & $ 82.55 \pm 1.47$ & $63.50 \pm 1.75$ & $81.85 \pm 0.74$ & $56.31\pm 2.20$ & $ 54.49 \pm 0.79$\\
LR & $55.64 \pm 1.89$ & $ 81.10 \pm 1.85$ & $55.21 \pm 2.21$ & $78.82 \pm 0.66$ & $ 55.25 \pm 1.48$ & $56.86 \pm 1.70$ \\
SRC & \textbf{67.66} $\pm$ \textbf{1.86} & $88.68 \pm 1.59$ & \textbf{69.29 $\pm$ 0.77} & \textbf{88.68 $\pm$ 1.03} & $ 65.43 \pm 1.24$ & \textbf{67.78 $\pm$ 1.76} \\
UDL & $64.68 \pm 2.11$ & $87.35 \pm 2.23$ & $67.35 \pm 1.22$ & $ 86.40 \pm 0.70$ & $ 64.36\pm 1.37$ & $ 65.23 \pm 2.02$ \\
SDL & $66.29\pm 1.81$ & \textbf{88.84 $\pm$ 2.31} & $68.61\pm 1.30$ & $ 87.50 \pm 0.82$ & \textbf{66.05 $\pm$ 0.75} & $67.31 \pm 1.38$ \\
\bottomrule
\end{tabular}
\end{table*} %¥

The WVU dataset consists of different biometric modalities such as fingerprint, iris, palmprint, hand geometry, and voice from subjects of different age, gender, and ethnicity. It is a challenging data set, as many of the samples are corrupted with blur, occlusion, and sensor noise. In this paper, two irises (left and right) and four fingerprint modalities are used. The evaluation is done on a subset of 202 subjects which have more than four samples in all modalities. Samples from different modalities are shown in Fig.~\ref{fig:SMDL}. The training set is formed by randomly selecting four samples from each subject, overall 808 samples. The remaining 509 samples are used for testing. The features used here are those described in~\cite{SPNC13} which are further PCA-transformed. The dimension of the input data after preprocessing are 178 and 550 for the fingerprint and iris modalities, respectively. All inputs are normalized to have zero mean and unit $l_2$ norm. The number of dictionary atoms for the dictionary learning algorithms are chosen to be 2 per class, resulting in dictionaries of overall 404 atoms. The dictionaries for JSRC and JDSRC are constructed using all the training samples. 

The classification results obtained using individual modalities on 5 different splits of the data into training and test samples are shown in Table~\ref{tab:WVUIndResults}. As shown, finger 2 is the strongest modality for the recognition task. % while finger 1 and 3 as well as the Irises have consistently weaker performances. 
The SRC and SDL algorithms achieve the best results. It should be noted that dictionary size of SRC is twice of that in SDL. 

\begin{figure}[t]
        \centering
                \includegraphics[width=0.45\textwidth]{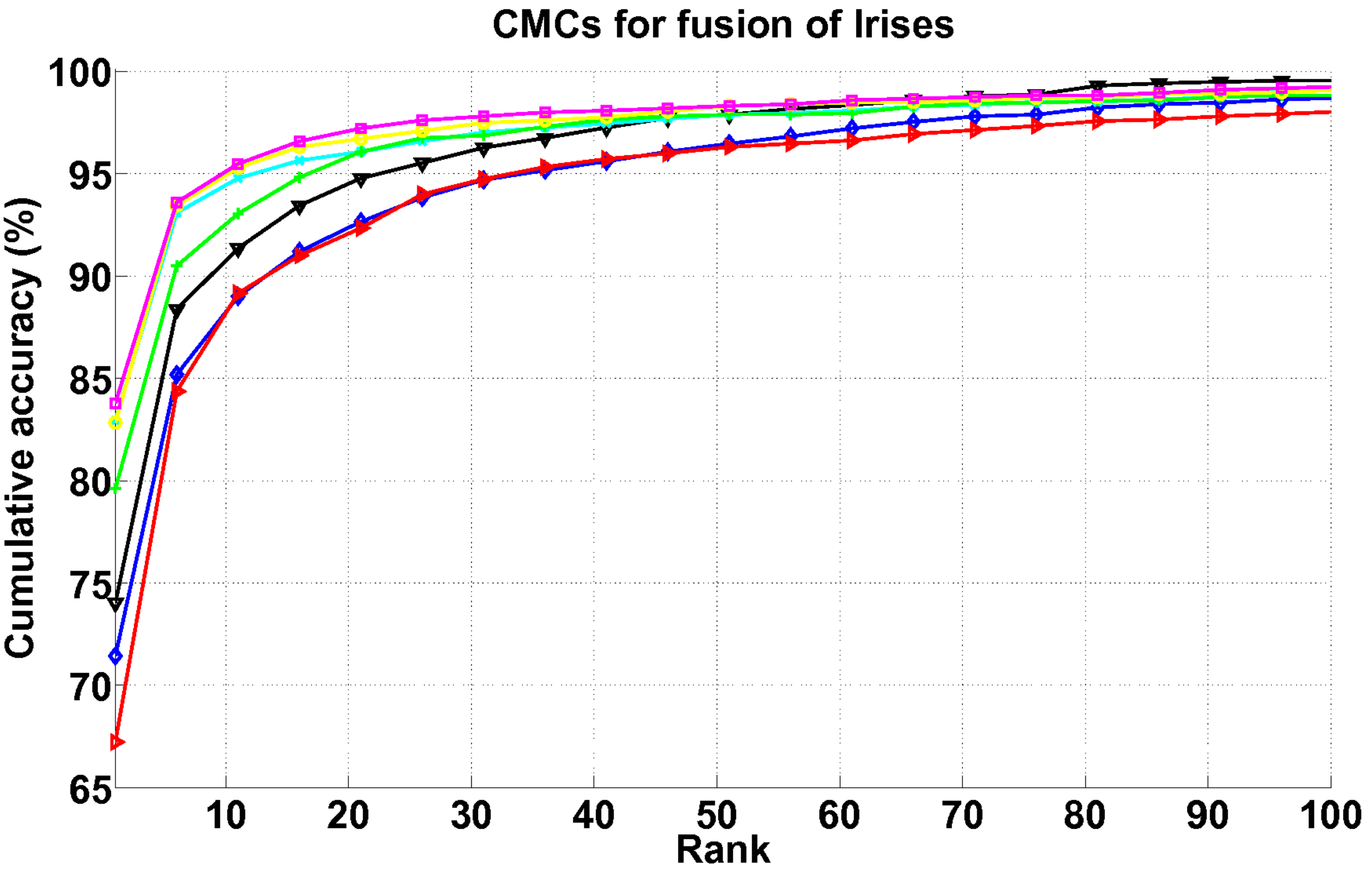}
                \includegraphics[width=0.45\textwidth]{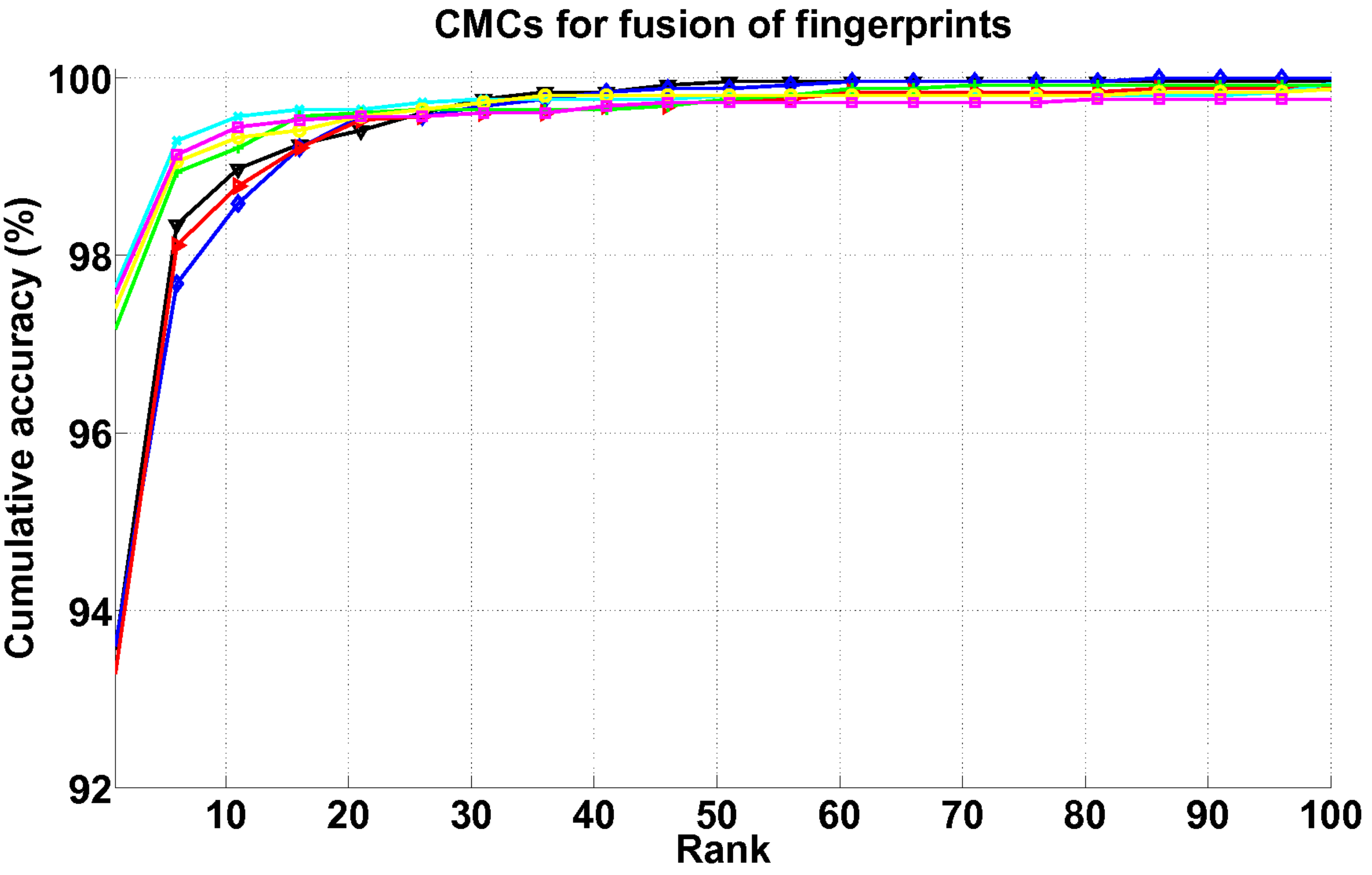}
                \includegraphics[width=0.45\textwidth]{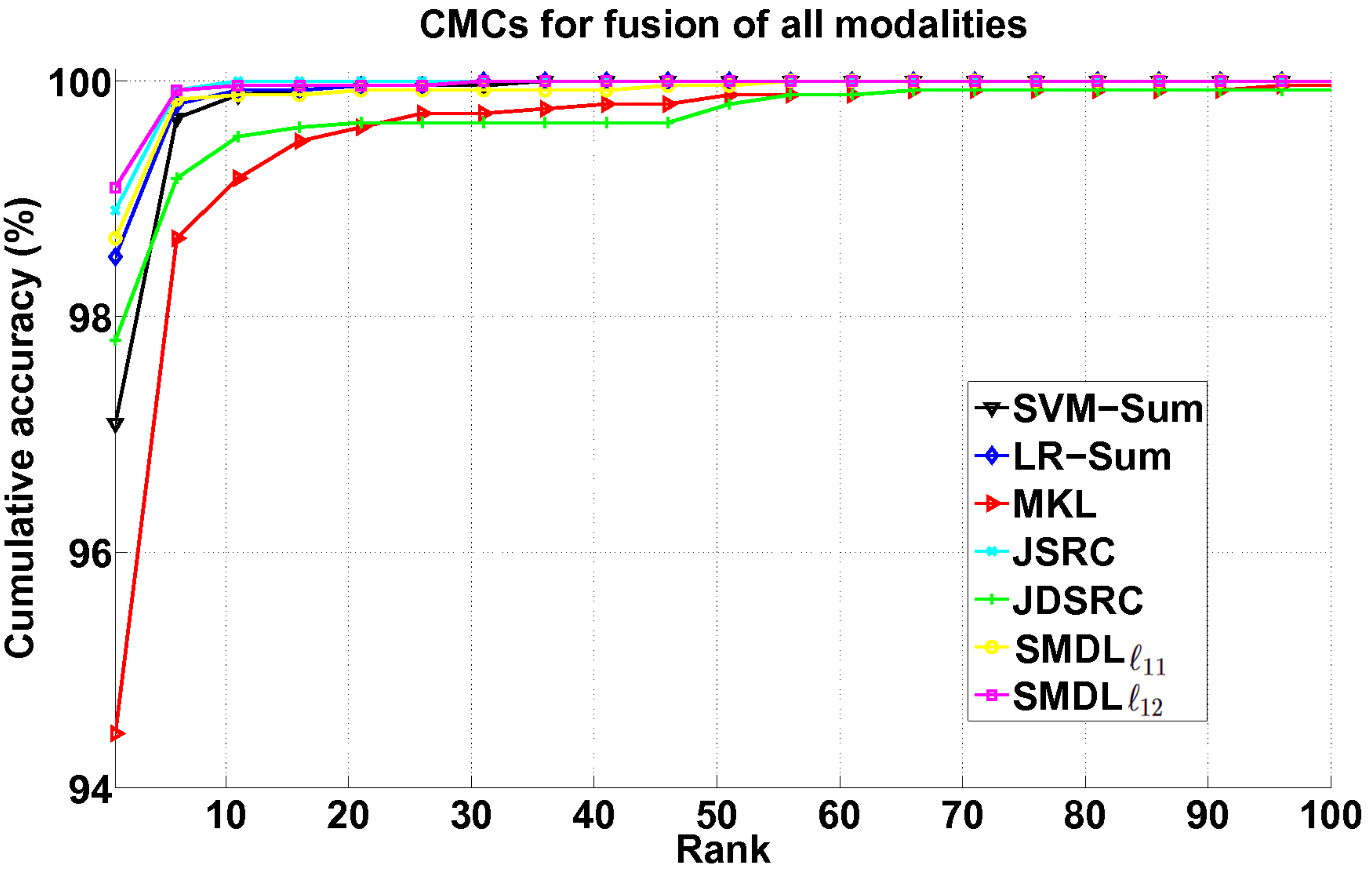}
        \caption{CMC plots obtained by fusing the Irises (top), fingerprints (middle), and all modalities (below) on the WVU dataset.}\label{fig:WVUCMCPlots}
\end{figure}

For multimodal classification, we consider fusion of fingerprints, fusion of Irises, and fusion of all the modalities. Table~\ref{tab:WVUFusResults} summarizes the correct classification rates of several fusion algorithms using 4 fingerprints, 2 Irises, and all the modalities, obtained on 5 different training and test splits. Fig.~\ref{fig:WVUCMCPlots} shows the corresponding cumulative matched score curves (CMC) for the competitive methods. CMC is a performance measure, similar to ROC, which is originally proposed for biometric recognition systems~\cite{BCPRS05}. As seen, the $\textrm{SMDL}_{\ell_{12}}$ algorithm outperforms the competitive algorithms and achieves the state-of-the-art performance using the Irises and all modalities with the rank one recognition rate of $83.77\%$ and $99.10\%$, respectively. Using the fingerprints, the performance of the $\textrm{SMDL}_{\ell_{12}}$ is close to the best performing algorithm, which is JSRC. The results suggest that using joint sparsity prior indeed improves the multimodal classification performance by extracting the coupled information among the modalities. %, especially when fusing the Iris modalities. This may be due to the strong correlation between the left and right irises compared to the fingerprints. %Adding an error term, as is done in~\cite{SPNC13}, can potentially improve the performance of the methods with joint sparsity priors by relaxing the joint sparse representation in the scenarios where the modalities are less correlated. However, this adds another regularization parameter which requires tuning and is avoided here. %This  limitation of joint sparse represenatation is discussed in~\cite{SPNC13, BRNJ14} in which an error term is added to allow for random sparsity, adding more flexibility to the model. 

\begin{figure}[!t]
   \centering
         \includegraphics[scale=0.17]{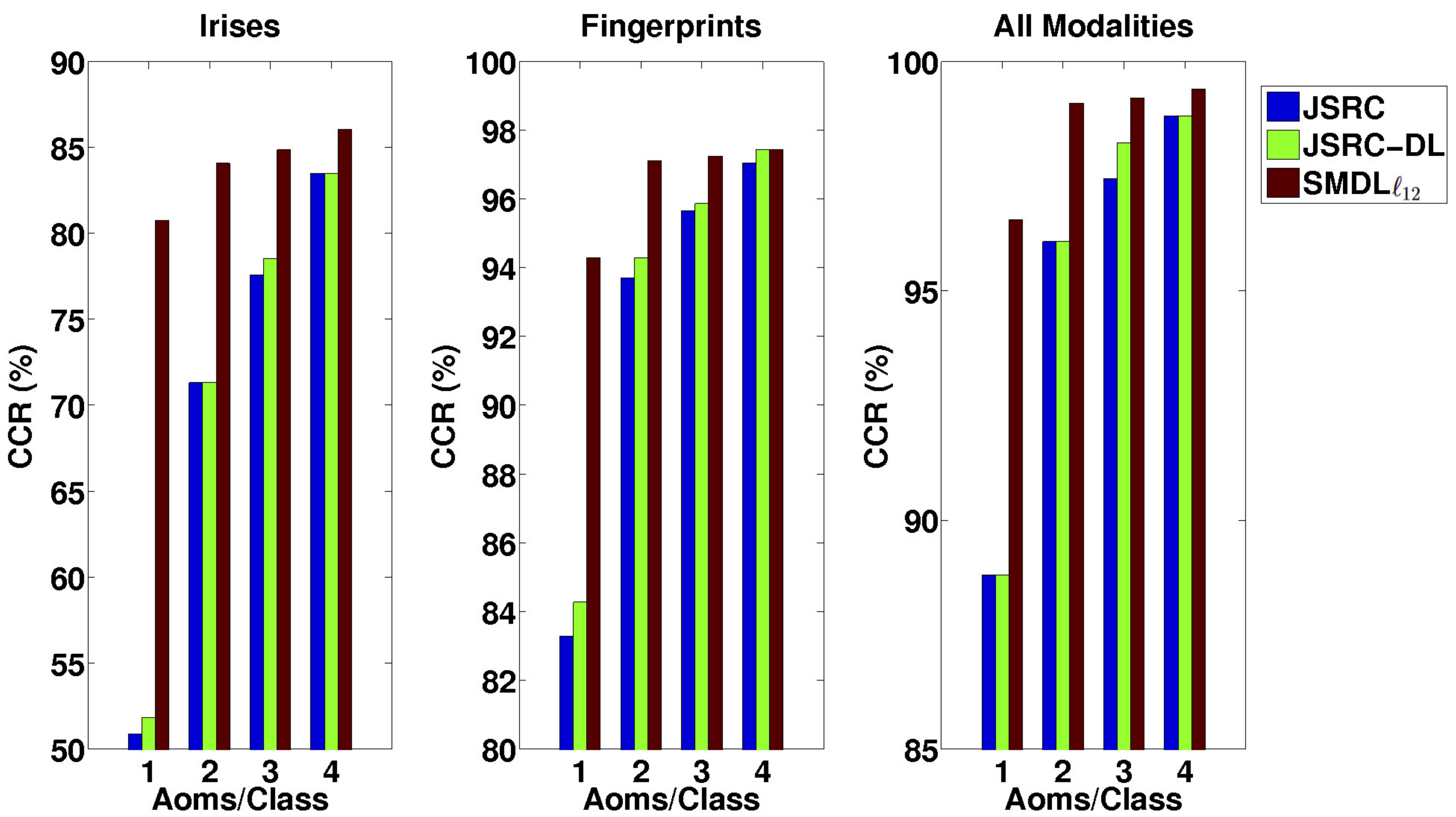}
         \caption{Comparison of the reconstructive-based (JSRC and JSRC-UDL) and the proposed discriminative-based ($\textrm{SMDL}_{\ell_{12}}$) classification algorithms obtained using the joint sparsity prior for different numbers of dictionary atoms per class on the WVU dataset.}
      \label{fig:VaryingAtom}
\end{figure}

Comparison of the algorithms with the joint sparsity priors indicates that the proposed $\textrm{SMDL}_{\ell_{12}}$ algorithm equipped with dictionaries of size 404 achieves comparable, and mostly better, results than the JSRC that uses dictionary of size 808. Similar to the experiment in Section~\ref{ssec:AR}, we compared the reconstructive and discriminating algorithms that are based on the joint sparsity prior when the number of dictionary atoms per class is kept equal. Fig.~\ref{fig:VaryingAtom} summarizes the results of the different fusion scenarios. As seen, $\textrm{SMDL}_{\ell_{12}}$ significantly outperforms JSRC and JSRC-UDL when the number of dictionary atoms per class is chosen to be 1 or 2. The results are consistent with that of Table~\ref{tab:VaryingAtomAR} for the AR dataset indicating that the proposed supervised formulation equipped with more compact dictionaries achieves superior performance than that of the reconstructive formulation for the studied biometric recognition applications.

\begin{table}[!t] %Table IV
\caption{Multimodal classification results obtained for the WVU dataset.}
\label{tab:WVUFusResults}
\centering
\small
\begin{tabular}{cccc}
Algorithm & 4 Fingerprints & 2 Irises & All modalities \\
\toprule
SVM-Maj & $90.14 \pm 0.70$ & $65.30 \pm 1.92$  & $95.24 \pm 0.92$ \\
SVM-Sum & $93.56 \pm 1.26$ & $74.03 \pm 1.89$ & $97.09 \pm 0.83$ \\
LR-Maj& $89.23 \pm 1.63$ & $63.73 \pm 1.29$& $94.18 \pm 1.13$ \\
LR-Sum & $93.60 \pm 0.96$ &$ 71.43 \pm 1.91$ & $98.51 \pm 0.18$  \\
MKL & $93.28 \pm 1.52$ &$67.23 \pm 0.70$ & $94.46 \pm  0.87$ \\
JSRC & \textbf{97.64 $\pm$ 0.44} & $82.94 \pm 0.78$ & $  98.89 \pm 0.30$ \\
JDSRC & $ 97.17 \pm 0.26$&$ 79.61 \pm 0.70$&$97.80\pm 0.51$ \\
$\textrm{UMDL}_{\ell_{11}}$ & $97.09 \pm 0.56$& $80.90\pm 0.61$& $ 98.62\pm 0.46$ \\
$\textrm{SMDL}_{\ell_{11}}$ & $97.41 \pm 0.71$  & $82.83 \pm 0.87$ &$ 98.66 \pm 0.43$ \\
$\textrm{UMDL}_{\ell_{12}}$ & $96.78 \pm 0.57 $& $81.53\pm 2.18$ & $ 98.78\pm 0.43$ \\
$\textrm{SMDL}_{\ell_{12}}$ & 97.56 $\pm$ 0.41 & \textbf{83.77 $\pm$ 0.89} & \textbf{99.10 $\pm$ 0.30} \\
\bottomrule
\end{tabular}
\end{table} %¥

%\begin{figure*}[t]
%        \centering
%		\begin{subfigure}[b]{0.43\textwidth}
%                \includegraphics[width=\textwidth]{Figures/CMCIrises.eps}
%                \caption{}
%        \end{subfigure}%
%        ~ %add desired spacing between images, e. g. ~, \quad, \qquad etc.
%          %(or a blank line to force the subfigure onto a new line)
%        \begin{subfigure}[b]{0.43\textwidth}
%                \includegraphics[width=\textwidth]{Figures/CMCFingerprints.eps}
%                \caption{}
%        \end{subfigure}
%        ~ %add desired spacing between images, e. g. ~, \quad, \qquad etc.
%          %(or a blank line to force the subfigure onto a new line)
%        \begin{subfigure}[b]{0.43\textwidth}
%                \includegraphics[width=\textwidth]{Figures/CMCAll.eps}
%                \caption{}
%        \end{subfigure}
%        \caption{CMC plots obtained by fusing the Irises (a), fingerprints (b), and all modalities on the WVU dataset.}\label{fig:WVUCMCPlots}
%\end{figure*}

\section{Conclusions and Future Works}\label{sec:Conclusions}
The problem of multimodal classification using sparsity models was studied and a task-driven formulation was proposed to jointly find the optimal dictionaries and classifiers under the joint sparsity prior. It was shown that the resulting bi-level optimization problem is smooth and an stochastic gradient descent algorithm was proposed to solve the corresponding optimization problem. The algorithm was then extended for a more general scenario where the sparsity prior was the combination of the joint and independent sparsity constraints. The simulation results on the studied image classification applications suggest that while the unsupervised dictionaries can be used for feature learning, the sparse coefficients generated by the proposed multimodal task-driven dictionary learning algorithms are usually more discriminative and therefore can result in improved multimodal classification performance. It was also shown that, compared to the  sparse-representation classification algorithms (JSRC, JDSRC, and JSRC-UDL), the proposed algorithms can achieve significantly better performance when compact dictionaries are utilized. 

In the proposed dictionary learning framework which utilizes the stochastic gradient algorithm, the learning rate should be carefully chosen for convergence of the algorithm. In out experiments, a heuristic was used to control the learning rate. Topics of future research include developing of better optimization tools for fast convergence guarantee in this non-convex setting. Moreover, developing task-driven dictionary learning algorithms under other proposed structured sparsity priors for multimodal fusion such as the tree-structured sparsity prior~\cite{BRNJ14, MJOB11} is another future research topic.
Future research will also include adapting of the proposed algorithms for other multimodal tasks such as multimodal retrieval, multimodal action recognition using Kinect data, and image super-resolution.

\appendix
The proof of Proposition~\ref{prop:MainProp} is presented using the following two results.

\lem[Optimality condition]
The matrix $\+A^{\star} = \left[ {\Balpha^1}^{\star} \dots {\Balpha^S}^{\star} \right] = \left[{\+a_{1\rightarrow}^{\star}}^T \dots {\+a_{d\rightarrow}^{\star}}^T \right]^T \in \mathbb{R}^{d \times S}$ is a minimizer of~(\ref{eq:JSRC2}) if and only if , $\forall j \in \lbrace 1, \dots, d\rbrace$,
%\begin{equation}\label{eq:OptimalityCond}
%\left\lbrace \begin{array}{ll} 
%\left[{\+d_j^1}^T\left(\+x^1 - \+D^1{\Balpha^1}^{\star}\right) \dots {\+d_j^S}^T\left(\+x^S - \+D^S{\Balpha^S}^{\star}\right) \right] -  \lambda_2 \+a_{j\rightarrow}^{\star} = \lambda_1\frac{\+a_{j\rightarrow}^{\star}}{\Vert \+a_{j\rightarrow}^{\star} \Vert_{\ell_2}} & \textrm{if} \Vert \+a_{j\rightarrow}^{\star} \Vert_{\ell_2} \neq 0, \\
%\Vert \left[{\+d_j^1}^T\left(\+x^1 - \+D^1{\Balpha^1}^{\star}\right) \dots {\+d_j^S}^T\left(\+x^S - \+D^S{\Balpha^S}^{\star}\right) \right] -  \lambda_2 \+a_{j\rightarrow}^{\star} \Vert_{\ell_2} \leq \lambda_1 & \textrm{otherwise}.
%\end{array}\right.
%\end{equation}
\begin{equation}\label{eq:OptimalityCond}
\left\lbrace \begin{array}{l}\begin{aligned}
&\left[{\+d_j^1}^T\left(\+x^1 - \+D^1{\Balpha^1}^{\star}\right) \dots {\+d_j^S}^T\left(\+x^S - \+D^S{\Balpha^S}^{\star}\right) \right] \\ 
&-\lambda_2 \+a_{j\rightarrow}^{\star} = \lambda_1\frac{\+a_{j\rightarrow}^{\star}}{\Vert \+a_{j\rightarrow}^{\star} \Vert_{\ell_2}}, \textrm{ if } \Vert \+a_{j\rightarrow}^{\star} \Vert_{\ell_2} \neq 0, \end{aligned}\\
\begin{aligned}
&\Vert \left[{\+d_j^1}^T\left(\+x^1 - \+D^1{\Balpha^1}^{\star}\right) \dots {\+d_j^S}^T\left(\+x^S - \+D^S{\Balpha^S}^{\star}\right) \right] \\
&-\lambda_2 \+a_{j\rightarrow}^{\star} \Vert_{\ell_2} \leq \lambda_1, \textrm{otherwise}.
\end{aligned}
\end{array}\right.
\end{equation}

\begin{proof}. The proof follows directly from the subgradient optimality condition of~(\ref{eq:JSRC2}), i.e.
%\[
%\+0 \in \left\lbrace \left[ -{\+D^1}^T\left(\+x^1 - \+D^1{\Balpha^1}^{\star}\right) \dots -{\+D^S}^T\left(\+x^S - \+D^S{\Balpha^S}^{\star}\right) \right] + \lambda_2\+A^{\star} + \lambda_1\+P : \+P \in \partial \Vert \+A^{\star} \Vert_{\ell_{12}} \right\rbrace,
%\]
\begin{equation*}
\begin{aligned}
\+0 \in \lbrace & \left[ {\+D^1}^T\left(\+D^1{\Balpha^1}^{\star}-\+x^1\right) \dots {\+D^S}^T\left(\+D^S{\Balpha^S}^{\star}-\+x^S\right) \right] \\ & + \lambda_2\+A^{\star} + \lambda_1\+P : \+P \in \partial \Vert \+A^{\star} \Vert_{\ell_{12}} \rbrace,
\end{aligned}
\end{equation*}
where $\partial \Vert \+A^{\star} \Vert_{\ell_{12}}$ denotes the subgradient of the $\ell_{12}$ norm evaluated at $\+A^{\star}$. As shown in~\cite{BF10}, the subgradient is characterized, for all $j \in \left\lbrace 1, \dots, d\right\rbrace$, as $\+p_{j\rightarrow} = \frac{\+a_{j\rightarrow}}{\Vert \+a_{j\rightarrow} \Vert_{\ell_2}}$ if $\Vert \+a_{j\rightarrow} \Vert_{\ell_2} > 0$, and $\Vert \+p_{j\rightarrow}\Vert_{\ell_2} \leq 1$ otherwise. 
\end{proof}

Before proceeding to the next proposition, we need to define the term \textit{transition point}. For a given $\lbrace\+x^s\rbrace$, let $\Lambda_{\lambda}$ be the active set of the solution $\+A^{\star}$ of~(\ref{eq:JSRC2}) when $\lambda_1 = \lambda$. Then $\lambda$ is defined to be a transition point of $\lbrace\+x^s\rbrace$ if $\Lambda_{\lambda + \epsilon} \neq \Lambda_{\lambda - \epsilon}, \forall \epsilon > 0$.

\prop[Regularity of $\+A^{\star}$]\label{prop:Regularity}
Let $\lambda_2 > 0$ and assumption (\textit{A}) be hold. Then,
 
 \textit{Part 1}. $\+A^{\star}(\lbrace\+x^s,\+D^s\rbrace )$ is a continuous function of $\lbrace\+x^s\rbrace$ and $\lbrace\+D^s\rbrace$.

\textit{Part 2}. If $\lambda_1$ is not a transition point of $\lbrace\+x^s\rbrace$, then the active set $\Lambda$ of $\+A^{\star}(\lbrace\+x^s,\+D^s\rbrace )$ is locally constant with respect to both $\lbrace\+x^s\rbrace$ and $\lbrace\+D^s\rbrace$. Moreover, $\+A^{\star}(\lbrace\+x^s,\+D^s\rbrace )$  is locally differentiable with respect to $\lbrace\+D^s\rbrace$. 

\textit{ Part 3}. $\forall \lambda_1 > 0, \exists$ a set $\mathcal{N}_{\lambda_1}$ of measure zero in which $\forall \lbrace\+x^s\rbrace \in \lbrace\mathbb{R}^{n_s}\rbrace \backslash \mathcal{N}_{\lambda_1}$, $\lambda_1$ is not any of the transition points of $\lbrace\+x^s\rbrace$. 

\begin{proof}.
Part 1. In the special case of $S=1$, which is equivalent to an elastic net problem, this has already been shown~\cite{MBPS09, ZHT07}. Our proof follows similar steps. Assumption (\textit{A}) guarantees that $\+A^{\star}$ is bounded. Therefore, we can restrict the optimization problem~(\ref{eq:JSRC2}) to a compact subset of $\mathbb{R}^{d \times S}$. Since $\+A^{\star}$ is unique (imposed by $\lambda_2 >0$) and the cost function of~(\ref{eq:JSRC2}) is continuous in $\+A$ and each element of the set $\lbrace\+x^s,\+D^s\rbrace$ is defined over a compact set, $\+A^{\star}(\lbrace\+x^s,\+D^s\rbrace )$ is a continuous function of $\lbrace\+x^s\rbrace$ and $\lbrace\+D^s\rbrace$.

Part 2 and Part 3. These statements are proved here by converting the optimization problem~(\ref{eq:JSRC2}) into an equivalent group lasso problem~\cite{YL06} and using some recent results on it. Let the matrix $\+{D}^\prime_j = \operatorname{blkdiag}(\+d^1_j, \dots, \+d^S_j) \in \mathbb{R}^{n \times S}, \forall j \in \lbrace 1, \dots, d\rbrace,$ be the block-diagnoal collection of the $j^{th}$ atoms of the dictionaries. Also let $\+{D}^\prime = \left[\+{D}^\prime_1 \dots \+{D}^\prime_d \right] \in \mathbb{R}^{n \times Sd}$, $\+{x}^\prime = \left[ {\+x^1}^T \dots  {\+x^S}^T \right]^T \in \mathbb{R}^{n}$, and $\+{a}^\prime = \left[ {\+a_{1\rightarrow}} \dots  {\+a_{d\rightarrow}} \right]^T \in \mathbb{R}^{Sd}$ . Then~(\ref{eq:JSRC2}) can be rewritten as 
\begin{equation}\label{eq:GroupLasso1}
\min_{\+A} \frac{1}{2}\Vert \+{x}^\prime - \+{D}^\prime\+{a}^\prime\Vert_{\ell_2}^2 +\lambda_1\sum_{j=1}^d \Vert \+a_{j\rightarrow} \Vert_{\ell_2} + \frac{\lambda_2}{2}\Vert \+A \Vert_{F}^2.
\end{equation}
This can be further converted into the standard group lasso:
\begin{equation}\label{eq:GroupLasso2}
\min_{\+A} \frac{1}{2}\Vert \+{x}^{\prime\prime} - \+{D}^{\prime\prime}\+{a}^\prime\Vert_{\ell_2}^2 +\lambda_1\sum_{j=1}^d \Vert \+a_{j\rightarrow} \Vert_{\ell_2},
\end{equation}
where $\+{x}^{\prime\prime} = \left[ {\+{x}^\prime}^T \+0^T\right]^T \in \mathbb{R}^{n+Sd}$ and $\+{D}^{\prime\prime} = \left[ {\+{D}^\prime}^T \sqrt{\lambda_2}\+I\right]^T \in \mathbb{R}^{(n+Sd) \times Sd}$. It is clear that the matrix $\+{D}^{\prime\prime}$ is full column rank. The rest of the proof follows directly from the results in~\cite{VDPFD12}. 
\end{proof}

\begin{proof}[Proof of Proposition~\ref{prop:MainProp}]
The above proposition implies that $\+A^{\star}$ is differentiable almost everywhere. We know prove the proposition~\ref{prop:MainProp}. It is easy to show that $f$ is differentiable with respect to $\+w^s$ due to the assumption (\textit{A}) and the fact that $l_{su}$ is twice differentiable. $f$ is also differentiable with respect to $\+D^s$ given assumption (\textit{A}), twice differentiability of  $l_{su}$, and the fact that $\+A^{\star}$ is differentiable everywhere except on a set of measure zero~ (Prop~\ref{prop:Regularity}). We obtain the derivative of $f$ with respect to $\+D^s$ using the chain rule. The steps are similar to those taken for $\ell_1$-related optimization in~\cite{YYH10}, though a bit more involved. Since the active set is locally constant, using the optimality condition~(\ref{eq:OptimalityCond}), we can implicitly differentiate $\+A^{\star}(\lbrace\+x^s,\+D^s\rbrace )$ with respect to $\+D^s$. For the non-active rows of $\+A^{\star}$, the differential is zero. On the active set $\Lambda$,~(\ref{eq:OptimalityCond}) can be rewritten as 
%\begin{equation}\label{eq:OptimalityCondAct}
%\left[{\+D_{\Lambda}^1}^T\left(\+x^1 - \+D^1{\Balpha^1}^{\star}\right) \dots {\+D_{\Lambda}^S}^T\left(\+x^S - \+D^S{\Balpha^S}^{\star}\right) \right] -  \lambda_2 \+A^{\star}_{\Lambda\rightarrow} = \lambda_1\left[\frac{{\+a_{1\rightarrow}^{\star}}^T}{\Vert \+a_{1\rightarrow}^{\star} \Vert_{\ell_2}} \dots \frac{{\+a_{N\rightarrow}^{\star}}^T}{\Vert \+a_{N\rightarrow}^{\star} \Vert_{\ell_2}} \right]^T,
%\end{equation}
\begin{equation}\label{eq:OptimalityCondAct}
\begin{aligned}
&\left[{\+D_{\Lambda}^1}^T\left(\+x^1 - \+D^1{\Balpha^1}^{\star}\right) \dots {\+D_{\Lambda}^S}^T\left(\+x^S - \+D^S{\Balpha^S}^{\star}\right) \right] \\ 
&-  \lambda_2 \+A^{\star}_{\Lambda\rightarrow} = \lambda_1\left[\frac{{\+a_{1\rightarrow}^{\star}}^T}{\Vert \+a_{1\rightarrow}^{\star} \Vert_{\ell_2}} \dots \frac{{\+a_{N\rightarrow}^{\star}}^T}{\Vert \+a_{N\rightarrow}^{\star} \Vert_{\ell_2}} \right]^T,
\end{aligned}
\end{equation}
where $N$ is the cardinality of $\Lambda$ and $\+D_{\Lambda}$ and $\+A^{\star}_{\Lambda\rightarrow}$ are the matrices consisting of active columns of $\+D$ and active rows of $\+A^{\star}$, respectively. For the rest of the proof, we only work on the active set and the symbols $\Lambda$ and $\star$ are dropped for the ease of notation. Taking the partial derivative from both sides of~(\ref{eq:OptimalityCondAct}) with respect to $d_{ij}^s$, the element in the $i^{th}$-row and $j^{th}$-column of $\+D^s$, and taking its transpose we have:
%\begin{equation}\label{eq:derivative1}
%\begin{aligned}
%\left[ \begin{array}{c}\+0 \\ \left(\+D^s{\Balpha^s}-\+x^s\right)^T\+E_{ij}^s + {\Balpha^s}^{T}{\+E_{ij}^s}^T\+D^s\\ \+0 \end{array}\right] +  &
%\left[ \begin{array}{c} \frac{\partial{\Balpha^1}^T}{\partial d_{ij}^s}{\+D^1}^T\+D^1 \\ \vdots \\ \frac{\partial{\Balpha^S}^T}{\partial d_{ij}^s}{\+D^S}^T\+D^S  \end{array}\right] + \lambda_2\frac{\partial{\+A}^T}{\partial d_{ij}^s} \\ & =  -\lambda_1\left[\+{\Delta}_1\frac{\partial\+a_{1\rightarrow}^T}{\partial d_{ij}^s} \dots \+{\Delta}_N\frac{\partial\+a_{N\rightarrow}^T}{\partial d_{ij}^s} \right],
%\end{aligned}
%\end{equation}
\begin{equation*}\label{eq:derivative1}
\begin{aligned}
& \lambda_2\frac{\partial{\+A}^T}{\partial d_{ij}^s} + \left[ \begin{array}{c}\+0 \\ \left(\+D^s{\Balpha^s}-\+x^s\right)^T\+E_{ij}^s + {\Balpha^s}^{T}{\+E_{ij}^s}^T\+D^s\\ \+0 \end{array}\right] +  \\
&\left[ \begin{array}{c} \frac{\partial{\Balpha^1}^T}{\partial d_{ij}^s}{\+D^1}^T\+D^1 \\ \vdots \\ \frac{\partial{\Balpha^S}^T}{\partial d_{ij}^s}{\+D^S}^T\+D^S  \end{array}\right]   =  -\lambda_1\left[\+{\Delta}_1\frac{\partial\+a_{1\rightarrow}^T}{\partial d_{ij}^s} \dots \+{\Delta}_N\frac{\partial\+a_{N\rightarrow}^T}{\partial d_{ij}^s} \right],
\end{aligned}
\end{equation*}
where $\+E_{ij}^s \in \mathbb{R}^{n_s \times N}$ is a matrix with zero elements except the element in the $i^{th}$ row and $j^{th}$ column which is one and 
\[
\+{\Delta}_k = \frac{1}{\Vert \+a_{k\rightarrow} \Vert_{\ell_2}}\left( I - \frac{1}{\Vert \+a_{k\rightarrow} \Vert_{\ell_2}^2 }\+a_{k\rightarrow}^T\+a_{k\rightarrow}\right) \in \mathbb{R}^{S\times S},\]
$\forall k \in \left\lbrace 1, \dots, N \right\rbrace$. It is easy to check that $\+{\Delta}_k \geq \+0$. Vectorizing the both sides and factorizing results in 
%\begin{equation}\label{eq:derivative2}
%\operatorname{vec}\left( \frac{\partial{\+A}^T}{\partial d_{ij}^s}\right) = \+P\left[ \begin{array}{c}\+0 \\ \left(\+x^s - \+D^s{\Balpha^s}\right)^T{\+e_{ij}^s}_1 - {\Balpha^s}^{T}{\+E_{ij}^s}^T\+d^s_1\\ \vdots \\ \left(\+x^s-\+D^s{\Balpha^s}\right)^T{\+e_{ij}^s}_N - {\Balpha^s}^{T}{\+E_{ij}^s}^T\+d^s_N \\ \+0 \end{array}\right],
%\end{equation}
\begin{equation}\label{eq:derivative2}
\begin{aligned}
& \operatorname{vec}\left( \frac{\partial{\+A}^T}{\partial d_{ij}^s}\right) = \\ 
&\+P\left[ \begin{array}{c}\+0 \\ \left(\+x^s - \+D^s{\Balpha^s}\right)^T{\+e_{ij}^s}_1 - {\Balpha^s}^{T}{\+E_{ij}^s}^T\+d^s_1\\ \vdots \\ \left(\+x^s-\+D^s{\Balpha^s}\right)^T{\+e_{ij}^s}_N - {\Balpha^s}^{T}{\+E_{ij}^s}^T\+d^s_N \\ \+0 \end{array}\right],
\end{aligned}
\end{equation}
where ${\+e_{ij}^s}_k$ is the $k^{th}$ column of $\+E_{ij}^s$, $\+P = \left( {\+{\hat{D}}}^T\+{\hat{D}} + \lambda_1\+{\Delta} + \lambda_2\+I\right)^{-1}$, and $\+{\hat{D}}$ and $\+{\Delta}$ are defined in Eqs.~(\ref{eq:Dhat}) and~(\ref{eq:Delta}), respectively. Further simplifying Eq.~(\ref{eq:derivative2}) yields
\begin{equation*}\label{eq:derivative3}
\operatorname{vec}\left( \frac{\partial{\+A}^T}{\partial d_{ij}^s}\right) = \+P_{\tilde{s}}{\+E_{ij}^s}^T\left(\+x^s - \+D^s{\Balpha^s}\right) - \+P_{\tilde{s}}{\+d_{i\rightarrow}^s}^{T}\alpha^s_j,
\end{equation*}
where $\tilde{s}$ is defined in Eq.~(\ref{eq:Prop1}). Using the chain rule, we have
\begin{equation*}\label{eq:derivative4}
 \frac{\partial{f}}{\partial d_{ij}^s} = \mathrm{E} \left[ \+g^T\operatorname{vec}\left( \frac{\partial{\+A}^T}{\partial d_{ij}^s}\right)\right], 
\end{equation*}
where $\+g = \operatorname{vec}\left(\frac{\partial \sum_{s=1}^S l_{su}}{\partial{\+A}^T}\right)$. Therefore, derivative with respective to the active columns of dictionary $\+D^s$ is 
\begin{equation*}\label{eq:derivative5}
\begin{aligned}
 &\frac{\partial{f}}{\partial \+D^s} =  \mathrm{E} \left[\begin{array}{c} \+g^T\+P_{\tilde{s}}\left({\+E_{11}^s}^T\left(\+x^s - \+D^s{\Balpha^s}\right)-{\+d_{1\rightarrow}^s}^{T}\alpha^s_1\right) \\ \vdots \\ \+g^T\+P_{\tilde{s}}\left( {\+E_{n_s1}^s}^T\left(\+x^s - \+D^s{\Balpha^s}\right)-{\+d_{n_s\rightarrow}^s}^{T}\alpha^s_1\right) 
 \end{array}\right. \\
& \left.\begin{array}{cc} \dots & \+g^T\+P_{\tilde{s}}\left({\+E_{1N}^s}^T\left(\+x^s - \+D^s{\Balpha^s}\right) -{\+d_{1\rightarrow}^s}^{T}\alpha^s_N\right)  \\ & \vdots \\ \dots & \+g^T\+P_{\tilde{s}}\left(  {\+E_{n_sN}^s}^T\left(\+x^s - \+D^s{\Balpha^s}\right)-{\+d_{n_s\rightarrow}^s}^{T}\alpha^s_N \right) 
 \end{array}\right] \\ &=  \mathrm{E} \left[\left(\+x^s - \+D^s{\Balpha^s}\right)\+g^T\+P_{\tilde{s}} - \+D^s\+P_{\tilde{s}}^T\+g{\alpha^s}^T\right].
\end{aligned}
\end{equation*}
Setting $\+{\beta} = \+P_{}^T\+g \in \mathbb{R}^{NS}$ and noting that $\+{\beta}_{\tilde{s}} =\+P_{\tilde{s}}^T\+g $ complete the proof. 
\end{proof}

Derivation of the algorithm with the mixed $\ell_{12}-\ell_{11}$ prior can be obtained similarly. For each active row $j \in \Lambda$ of $\+A^{\star}$, the solution of the optimization problem~(\ref{eq:JSRC+l1}) with the mixed prior, let $\Pi_j \subseteq \mathcal{S}$ be the set of active modalities which have non-zeros entries. Then the optimality condition for the active row $j$ is
\begin{equation*}
\begin{aligned}
&\left[{\+d_j^1}^T\left(\+x^1 - \+D^1{\Balpha^1}^{\star}\right) \dots {\+d_j^S}^T\left(\+x^S - \+D^S{\Balpha^S}^{\star}\right) \right]_{\Pi_j } \\ 
&-\lambda_2 \+a_{j\rightarrow}^{\star} = \lambda_1\frac{\+a_{j\rightarrow,\Pi_j}^{\star}}{\Vert \+a_{j\rightarrow}^{\star} \Vert_{\ell_2}} + \lambda_1^{\prime}\operatorname{sign}\left(\+a_{j\rightarrow,\Pi_j}^{\star} \right). 
\end{aligned}
\end{equation*}
Then, the algorithm for the mixed prior can be obtained by differentiating the optimality condition, following similar steps as was shown for the $\ell_{12}$ prior.

% -------------------------------------------------------------------------
\bibliographystyle{IEEEbib}
\bibliography{multiModalTaskDrivenFinVer}

\begin{IEEEbiography}[{\includegraphics[width=1in,height=1.25in,clip,keepaspectratio]{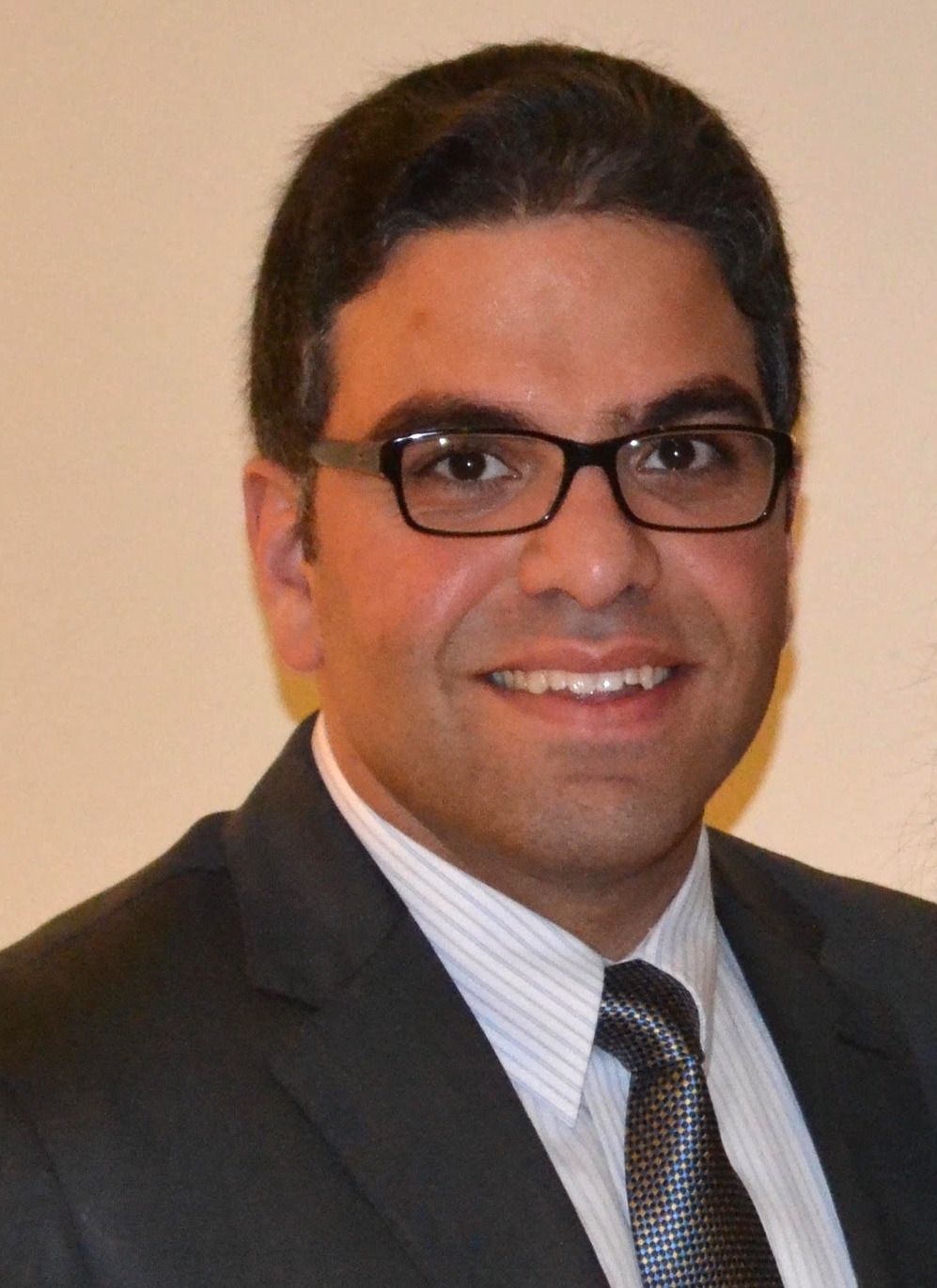}}]%
{Soheil Bahrampour} received the M.Sc. degree in electrical engineering from the University of Tehran, Iran, in 2009. He then received the M.Sc. degree in Mechanical Engineering and PhD degree in Electrical Engineering under the supervision of A. Ray and W.K. Jenkins from The Pennsylvania State University, University Park, PA, in 2013 and 2015, respectively. Dr. Bahrampour is currently a Research Scientist with Bosch Research and Technology Center, Palo Alto, CA. His research interests include machine learning, data mining, signal processing, and computer vision.\end{IEEEbiography}

\begin{IEEEbiography}[{\includegraphics[width=1in,height=1.25in,clip,keepaspectratio]{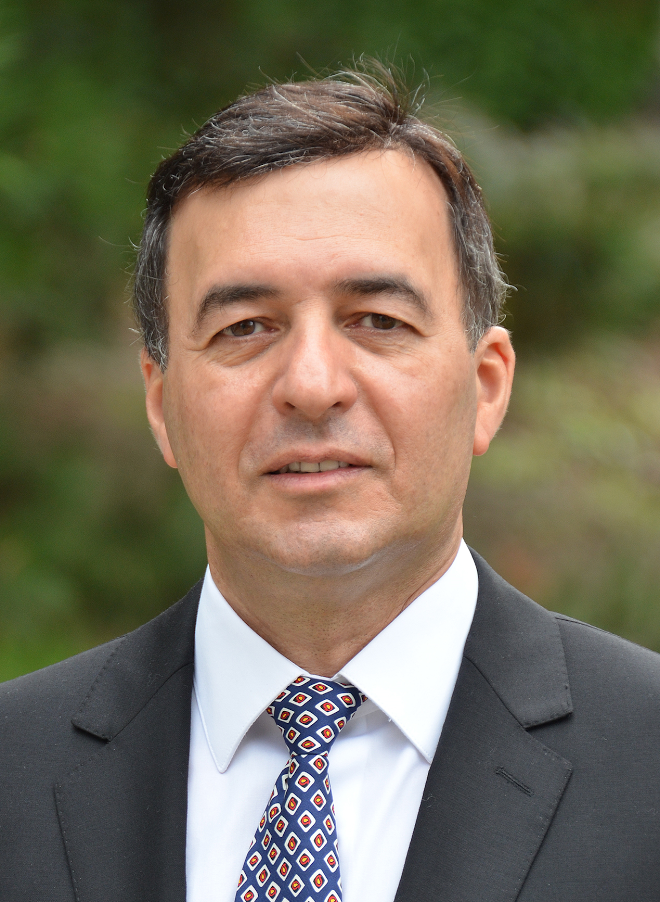}}]%
{Nasser M. Nasrabadi} (S'80-M'84-SM'92-FM'01) received the B.Sc. (Eng.) and Ph.D. degrees in Electrical Engineering from Imperial College of Science and Technology (University of London), London, England, in 1980 and 1984, respectively.
From October 1984 to December 1984 he worked for IBM (UK) as a senior programmer. During 1985 to 1986 he worked with Philips research laboratory in NY as a member of technical staff. From 1986 to 1991 he was an assistant professor in the Department of Electrical Engineering at Worcester Polytechnic Institute, Worcester, MA. From 1991 to 1996 he was an associate professor with the Department of Electrical and Computer Engineering at State University of New York at Buffalo, Buffalo, NY.  Since September From 1996 to 2015 he was a Senior Research Scientist (ST) with the US Army Research Laboratory (ARL).  Since August 2015 he has been a professor at Lane Dept. of Computer Science and Elecrical Engineering.  Dr. Nasrabadi has served as an associate editor for the IEEE Transactions on Image Processing, the IEEE Transactions on Circuits, Systems and Video Technology, and the IEEE Transactions on Neural Networks. His current research interests are in image processing, computer vision, biometrics, statistical machine learning theory, sparsity, robotics, and neural networks applications to image processing. He is also a Fellow of ARL, SPIE and IEEE.
\end{IEEEbiography}

\begin{IEEEbiography}[{\includegraphics[width=1in,height=1.25in,clip,keepaspectratio]{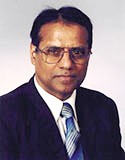}}]%
{Asok Ray} (SM'83–F'02) received the Ph.D. degree in Mechanical Engineering from Northeastern University, Boston, MA,  and the graduate degrees in the disciplines of Electrical Engineering, Mathematics, and Computer Science. He joined The Pennsylvania State University (Penn State), University Park, PA, in July 1985, and is currently a Distinguished Professor of Mechanical Engineering and Mathematics, a Graduate Faculty of Electrical Engineering, and a Graduate Faculty of Nuclear Engineering.  Prior  to  joining  Penn  State, he  held  research  and  academic  positions  with Massachusetts Institute of Technology, Cambridge, MA, and Carnegie-Mellon University,  Pittsburgh,  PA,  as  well  as  management  and research positions  with GTE Strategic Systems Division, Westborough, MA, Charles Stark Draper Laboratory, Cambridge, MA, and MITRE Corporation, Bedford, MA.  Dr. Ray has  authored  or  coauthored  over  550  research  publications, including  over  285  scholarly  articles  in  refereed  journals  and  research monographs. Dr. Ray is also a Fellow of the American Society of Mechanical Engineers (ASME) and a Fellow of World Innovative Foundation (WIF).  Dr. Ray had been a Senior Research Fellow at NASA Glenn Research Center under a National Academy of Sciences award.\end{IEEEbiography}

\begin{IEEEbiography}[{\includegraphics[width=1in,height=1.25in,clip,keepaspectratio]{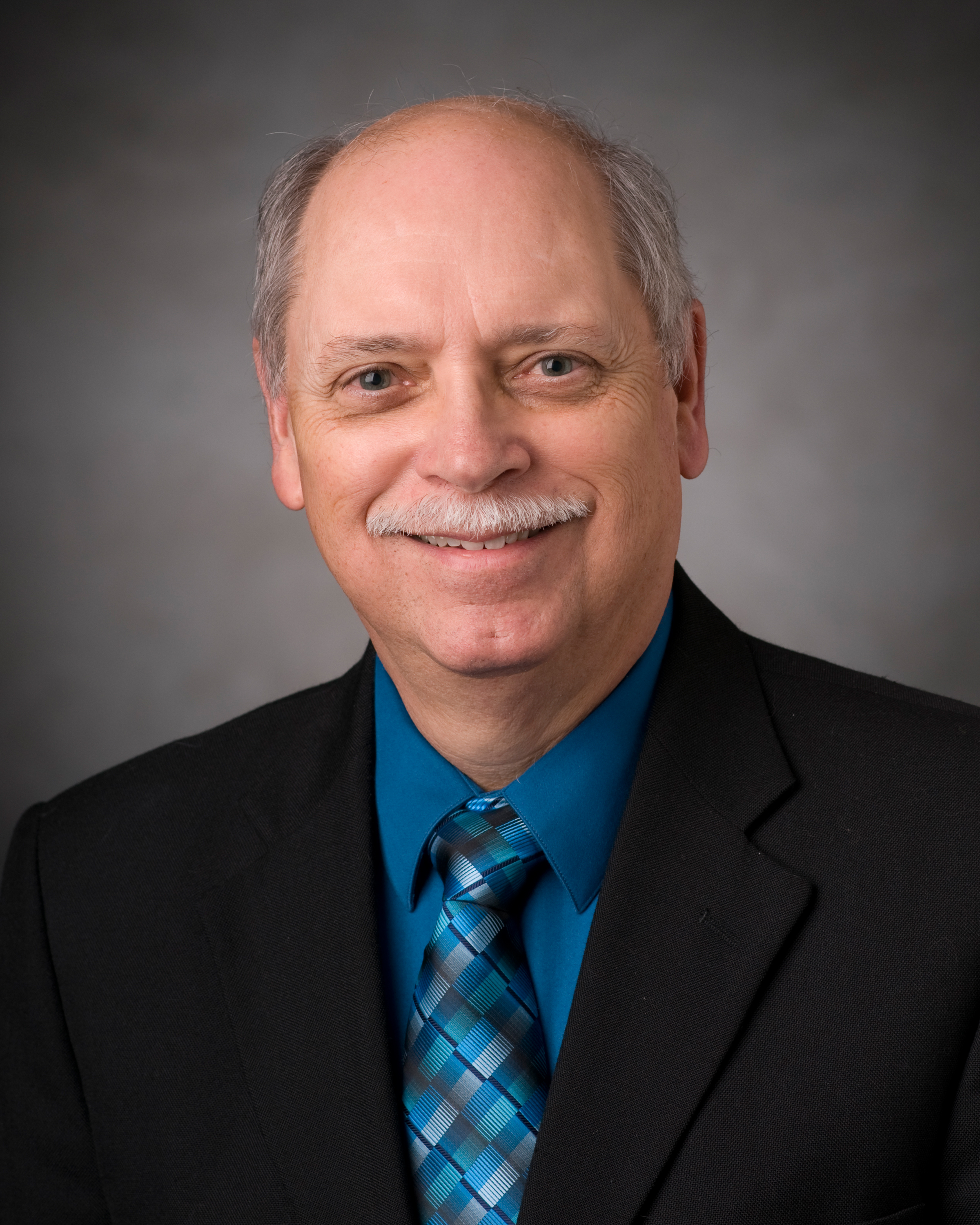}}]%
{W. Kenneth Jenkins} received the B.S.E.E. degree from Lehigh University and the M.S.E.E. and Ph.D. degrees from Purdue University. From 1974 to 1977 he was a Research Scientist Associate in the Communication Sciences Laboratory at the Lockheed Research Laboratory, Palo Alto, CA. In 1977 he joined the University of Illinois at Urbana-Champaign where he was a faculty member in Electrical and Computer Engineering from 1977 until 1999. From 1986-1999 Dr. Jenkins was the Director of the Coordinated Science Laboratory. From 1999 through 2011 he served Professor and Head of Electrical Engineering at Penn State University, and in 2011 he returned to the rank of Professor of Electrical Engineering.
Dr. Jenkins’ current research interests include fault tolerant DSP for highly scaled VLSI systems, adaptive signal processing, multidimensional array processing, computer imaging, bio-inspired optimization algorithms for intelligent signal processing, and fault tolerant digital signal processing. He co-authored the book Advanced Concepts in Adaptive Signal Processing, published by Kluwer in 1996. He is a past Associate Editor for the IEEE Transaction on Circuits and Systems, and a past President (1985) of the CAS Society. He served as General Chairman of the 1988 Midwest Symposium on Circuits and Systems and as the General Chairman of the Thirty Second Annual Asilomar Conference on Signals and Systems. From 2002 to 2007 he served on the Board of Directors of the Electrical and Computer Engineering Department Heads Association (ECEDHA) and as President of ECEDHA in 2005. Since January 2011 he has been serving as a Member of the IEEE-HKN Board of Governors. 
Dr. Jenkins is a Life Fellow of the IEEE and a recipient of the 1990 Distinguished Service Award of the IEEE Circuits and Systems Society. In 2000 he received a Golden Jubilee Medal from the IEEE Circuits and Systems Society and a 2000 Millennium Award from the IEEE. In 2000 was named a co-winner of the 2000 International Award of theGeorge Montefiore Foundation (Belgium) for outstanding career contributions to the field of electrical engineering and electrical science, in 2002 he was awarded the Shaler Area High School Distinguished Alumnus Award, in 2007 he was honored with an IEEE Midwest Symposium on Circuits and Systems 50th Anniversary Award, and in 2013 he received the ECEDHA Robert M. Janowiak Outstanding Leadership and Service Award. \end{IEEEbiography}

\end{document}